# Sarrus-inspired Deployable Polyhedral Mechanisms

Yuanqing Gu [a], Xiao Zhang [a], Guowu Wei [b,*], Yan Chen [a,c,†]

[a] School of Mechanical Engineering, Tianjin University, Tianjin 300350, China

[b] School of Science, Engineering and Environment, University of Salford, Salford M5 4WT, UK

[c] Key Laboratory of Mechanism Theory and Equipment Design of Ministry of Education, Tianjin University, 135 Yaguan Road, Tianjin 300350, China

**Abstract**

Deployable polyhedral mechanisms (DPMs) have witnessed flourishing growth in recent years because of their potential applications in robotics, space exploration, structure engineering, etc. This paper firstly presents the construction, mobility and kinematics of a family of Sarrus-inspired deployable polyhedral mechanisms. By carrying out expansion operation and implanting Sarrus linkages along the straight-line motion paths, deployable tetrahedral, cubic and dodecahedral mechanisms are identified and constructed following tetrahedral, octahedral and icosahedral symmetry, respectively. Three paired transformations with synchronized radial motion between Platonic and Archimedean polyhedrons are revealed, and their significant symmetric properties are perfectly remained in each work configuration. Subsequently, with assistant of equivalent prismatic joints, the equivalent analysis strategy for mobility of multiloop polyhedral mechanisms is proposed to significantly simplify the calculation process. This paper hence presents the construction method and equivalent analysis of the Sarrus-inspired DPMs that are not only valuable in theoretical investigation, but also have great potential in practical applications such as mechanical metamaterials, deployable architectures and space exploration.

## 1 Introduction

Deployable polyhedral mechanisms (DPMs) [1] have aroused great interest from researchers in the fields of robotics, space exploration, structure engineering and so on. After Verheyen [2] reported the pioneering work for the expandable polyhedral structures known as Jitterbug transformers [3], the construction methods of Fulleroid-like polyhedral mechanisms [4] were developed by Wohlhart [5], Kiper [6] and Röschel [7]. Furthermore, several overconstrained linkages [8] were adopted for the synthesis of DPMs. Kiper and Söylemez [9] introduced deployable polyhedrons by integrating multiple loops of equilateral Bennett linkages [10] with a rather small expansion ratio. Nevertheless, based on the Bennett and Bricard linkages [11], Yang et al. [12-14] realized three one-DOF transformations with a large volumetric ratio between Platonic and Archimedean solids, respectively. Wang and Kong [15-16] demonstrated a family of overconstrained multi-loop DPMs by connecting orthogonal single-loop linkages, including the Bricard linkage, using *S*-joints. Xiu et al. [17] developed a synthesis approach for generating Fulleroid-like Platonic and Archimedean DPMs based on the Sarrus-like overconstrained eight-bar linkages. Moreover, various DPMs based on prisms and antiprisms have been developed [18-22]. It can be found that, due to a large number of redundant constraints, most of them are multi-loop overconstrained mechanisms [23].

* Corresponding author. *E-mail address*: g.wei@salford.ac.uk (G. Wei)

† Corresponding author. *E-mail address*: yan_chen@tju.edu.cn (Y. Chen)

Apart from the above investigations, there is one type of DPMs that is capable of performing radial motions. A popular toy named Hoberman Sphere [24] with one-DOF radial motion was produced by combining Sarrus linkages [25] and scissor-like elements. To retain the exterior shape during deployment, Agrawal et al. [26] set up the radially expanding polyhedrons by introducing prismatic joints to polyhedral edges. In addition, Wei et al. [27-29] proposed a synthesis mothed of plane-symmetric eight-bar linkages for constructing regular and semi-regular DPMs possessing one-degree-of-freedom (DOF) radially reciprocating motion. Furthermore, reconfigurable DPMs constructed by using a variable revolute joint were introduced by Wei [30] in 2014. Similar polyhedral mechanisms were synthesized by integrating the assembly of planar mechanisms into the edges [31-32] and facets [33-36] of various polyhedrons. Recently, taking the one-DOF polygonal prisms as basic units, a group DPMs were proposed based on an additive-then-subtractive design strategy [37].

Here, we aim to design DPMs based on the transformations between Platonic and Archimedean polyhedrons, as shown in Fig. 1, referring to the polyhedral expansion in solid geometry [38]. Taking the case shown in Fig. 1(a) as an example, during the transformation, the blue facets in a Platonic polyhedron are separated and moved radially apart, and new facets in beige are formed among separated elements to form a corresponding Archimedean polyhedron. Due to the expansion operation, two polyhedrons in each transformation have identical polyhedral symmetry property, from Figs.1(a)-(c) are tetrahedral symmetry ($T_d$), octahedral symmetry ($O_h$) and icosahedral symmetry ($I_h$), respectively. Yet, it is extremely challenging to accomplish such transformation from mechanism point of view. Therefore, this paper aims at presenting a novel synthesis method for constructing a group of Sarrus-inspired DPMs with symmetric transformability. As the first published overconstrained linkage, the Sarrus linkage is adopted as the construction unit in this paper, which can generate exact straight-line motion. For demonstration purpose, we first construct the deployable tetrahedral mechanism that can transform a tetrahedron into a rhombitetratetrahedron, and verse visa, and then extend the method to the construction of deployable cubic and dodecahedral mechanisms. The equivalent analysis is also presented, followed by the mobility and kinematic analysis.

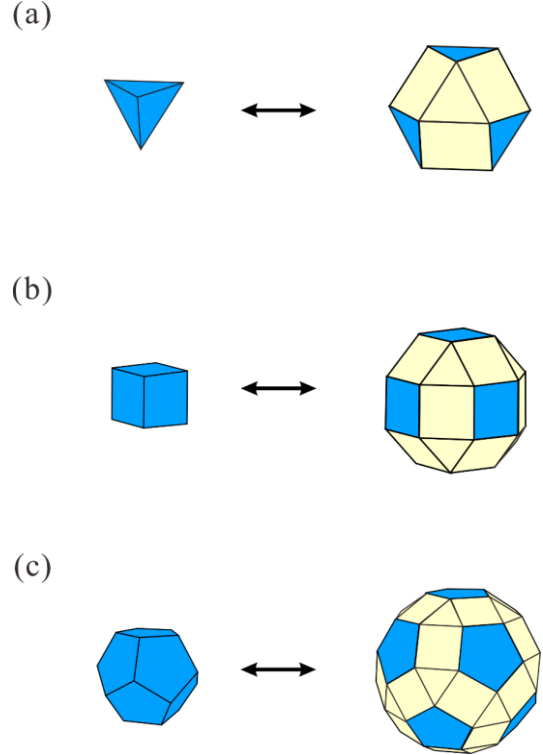

**Fig. 1.** Three paired Platonic and Archimedean polyhedrons. (a) A tetrahedron and a rhombitetratetrahedron with tetrahedral symmetry ($T_d$); (b) a cube and a rhombicuboctahedron with octahedral symmetry ($O_h$); (c) a dodecahedron and a rhombicosidodecahedron with icosahedral symmetry ($I_h$).

**2 A Sarrus-inspired deployable tetrahedral mechanism**

This section presents the construction of a deployable tetrahedral mechanism by integrating Sarrus linkages into a radially decomposed tetrahedron. The procedures for generating the proposed mechanism are introduced and the mobility analysis is conducted and verified through the reciprocal derivation.

**2.1 Construction of a deployable tetrahedral mechanism**

A hollow tetrahedron with four congruous prismoid platforms A, B, C and D is given in Fig. 2(a). Its four vertices are denoted by $a$ to $d$, six edges are $ab$, $ac$, $ad$, $bc$, $cd$ and $bd$ with the $x$-axis passing through midpoints of edges $ad$ and $bc$, so do the $y$-axis for edges $ac$ and $bd$, and $z$-axis for edges $ab$ and $cd$. Here, the coordinate origin O is the centroid of this tetrahedron, and the perpendiculars of the four platforms intersecting at the centroid O are denoted by red dash-dot lines. Subsequently, by carrying out the expansion operation, four platforms are separated synchronously and moved radially along the corresponding perpendiculars, and each two adjacent triangular prismoid platforms a straight-line motion, see Fig. 2(b). Based on tetrahedral geometry, the angle between the bottom and side facets of each triangular prismoid is $\beta=35.26°$, i.e., a half of the dihedral angle (70.53°) between prismoids A and B. Moreover, any edge of the tetrahedron is divided into two edges, such as $ab$ into $a_1b_1$ and $a_2b_2$. At this moment, the trends of straight-line motion occur between any two adjacent platforms whilst they are away from the centroid, which are represented by red solid lines. For instance, virtual straight-line motion path $p_1$ between platforms A and B is parallel to the line $a_1a_2$, or $b_1b_2$. In order to enable this motion, the Sarrus linkage is adopted in this paper as it can generate the exact straight-line motion between two platforms. As shown in Fig. 2(c), one Sarrus linkage between platforms A and B consists of six rigid bodies connected by six revolute joints, three parallel joints with axes $z_1$, $z_2$ and $z_3$ are implanted along line $a_1a_2$, so do the other three joints with axes $z_4$, $z_5$ and $z_6$ along $b_1b_2$. Here, the angle between the revolute axes in two limbs in a Sarrus linkage is $\gamma$, to avoid physical interference in the fully folded configuration and consider the tetrahedral geometry, $\gamma \in (0, 70.53°]$ should be satisfied. Meanwhile, due to the radial decomposition of triangular prismoids, for instance, side facet of platform A is coplanar with a virtual plane of axes $z_3$ and $z_6$. Thus, the straight-line motion along $p_1$ between platforms A and B is obtained.

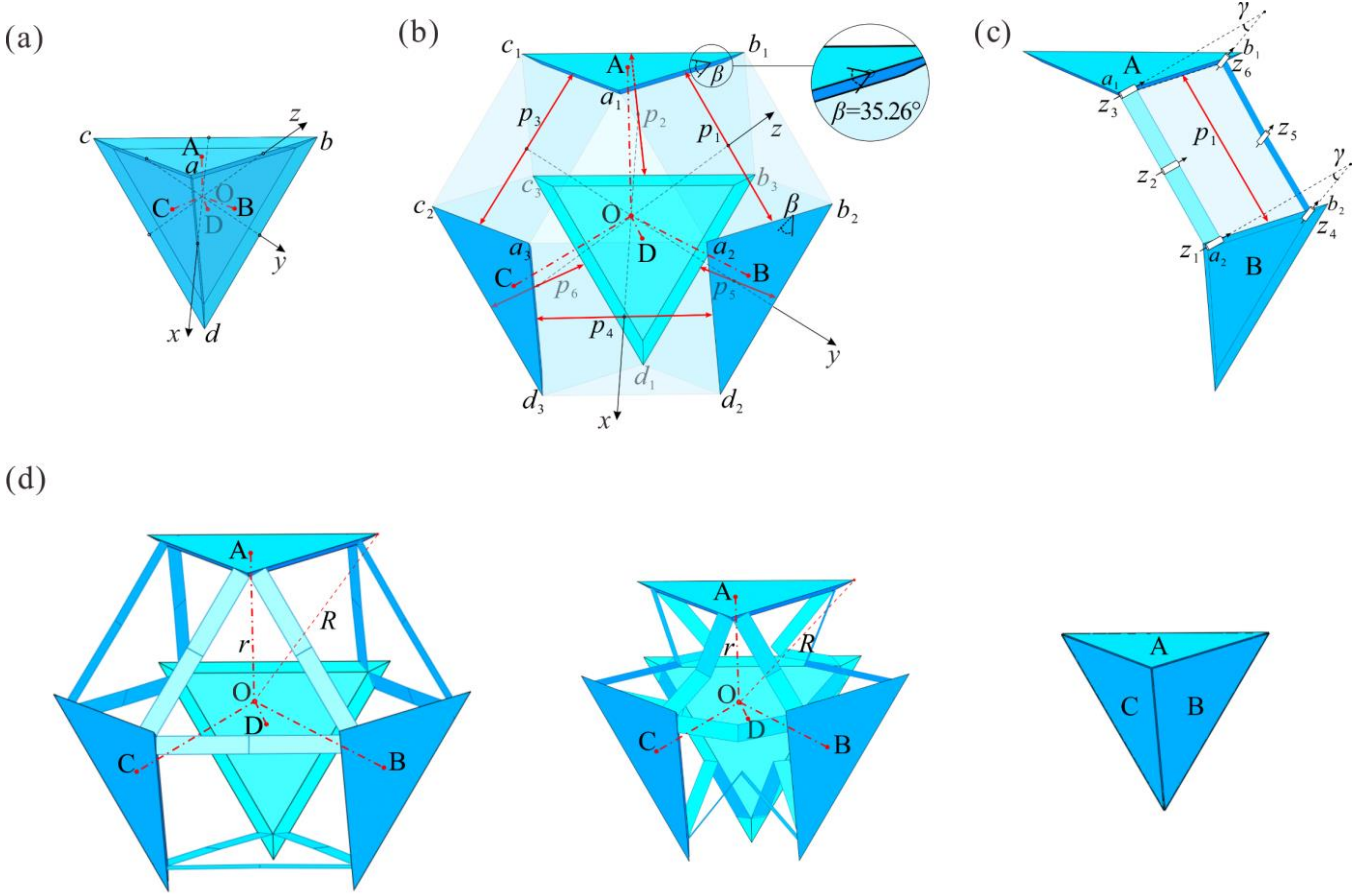

**Fig. 2.** Construction of a deployable tetrahedral mechanism. (a) A tetrahedron and the Cartesian coordf4inate system; (b) the expansion of four triangular platforms of a tetrahedron; (c) the Sarrus linkage with platforms A and B; (d) the motion sequence (transformation) from a rhombitetratetrahedron (the deployed configuration) to a tetrahedron (the folded configuration).

Furthermore, we can take the similar implantation and integrate five extra Sarrus linkages into each pair of two adjacent platforms along paths $p_2$ to $p_6$ following the procedure along $p_1$, in which the geometric relations of all integrated Sarrus linkages are identical. Thus, a novel deployable tetrahedral mechanism is obtained in Fig. 2(d), which consequently leads to a transformation from a rhombitetratetrahedron (the deployed configuration) to a tetrahedron (the folded configuration). Moreover, we can make $\gamma_{max} = 70.53°$ to obtain a planar triangle in a fully deployed configuration among three different platform vertices (such as $a_1, a_2$ and $a_3$), which are composed of three limbs of three adjacent Sarrus linkages. Based on such a construction, it can be seen that the Sarrus-inspired deployable tetrahedron performs synchronized radial motion. The four platforms have straight-line motion along their respective perpendiculars relative to the centroid O while separating from each other, which mechanically presents the expansion operation in geometry. It should be pointed out that the four platforms A to D locate on the faces of a virtual tetrahedron during the continuous motion process, in which the $T_d$ symmetry of this deployable tetrahedron is completely reserved. This mechanism has one DOF which can be theoretically verified in the following section.

**2.2 Mobility analysis and the reciprocal verification**

Mobility of the proposed deployable tetrahedral mechanisms can be investigated with screw theory [39]. First, as the foundational element of polyhedral construction, a Sarrus linkage in an arbitrary configuration is given in Fig. 3(a) attached with a local coordinate frame $\{x_i, y_i, z_i\}$, where the origin $O_i$ locates at the centre of virtual plane between two mobile platforms, $y_i$-axis is aligned with the straight-line motion direction and $z_i$-axis is perpendicular to the virtual plane. It is well known that this linkage consists of two limbs, motion-screw system of limb 1 can be calculated in the associated local coordinate system as

$$\mathbb{S}_{l1} = \begin{cases} S_{i1} = [\sin\alpha \quad 0 \quad \cos\alpha \quad -b\cos\alpha\sin\varphi \quad l\cos\alpha \quad b\sin\alpha\sin\varphi]^T \\ S_{i2} = [\sin\alpha \quad 0 \quad \cos\alpha \quad 0 \quad l\cos\alpha - b\cos\varphi \quad 0]^T \\ S_{i3} = [\sin\alpha \quad 0 \quad \cos\alpha \quad b\cos\alpha\sin\varphi \quad l\cos\alpha \quad -b\sin\alpha\sin\varphi]^T \end{cases}, \quad (1)$$

and for limb 2,

$$\mathbb{S}_{l2} = \begin{cases} S_{i4} = [-\sin\alpha \quad 0 \quad \cos\alpha \quad -b\cos\alpha\sin\varphi \quad -l\cos\alpha \quad -b\sin\alpha\sin\varphi]^T \\ S_{i5} = [-\sin\alpha \quad 0 \quad \cos\alpha \quad 0 \quad b\cos\varphi - l\cos\alpha \quad 0]^T \\ S_{i6} = [-\sin\alpha \quad 0 \quad \cos\alpha \quad b\cos\alpha\sin\varphi \quad -l\cos\alpha \quad b\sin\alpha\sin\varphi]^T \end{cases}, \quad (2)$$

where subscript $i$ indicates the number of Sarrus linkages involved in the integration of the proposed deployable polyhedral mechanisms, $\varphi$ is the folding angle between a half of limb and platform, $b$ is half of the edge length $a$ of a regular rhombitetratetrahedron, as well as $l = a/2$ and $\alpha = \gamma/2$.

The reference coordinate frame $\{x, y, z\}$ in a deployed configuration of the tetrahedral mechanism is established in Fig. 3(b), in which its reference origin O locates at the centroid of the rhombitetratetrahedron and six local origins $O_i$ ($i$ =1, 2, …, 6) are the centres of virtual planes expanded by any two adjacent platforms, respectively. Conceivably, the four platforms A, B, C and D are throughout located at the vertices of a virtual dual tetrahedron during the continuous motion process, thus a dual tetrahedron is introduced in Fig. 3(c) to conveniently describe the directions of axes $y_i$, i.e. the direction of straight-line motion between two adjacent platforms. The motion screws of a Sarrus linkage element in Fig. 3(a) can be transformed to the reference coordinate system in Fig. 3(b) with the adjoint transformation matrix $Ad_T = \begin{bmatrix} \mathbf{R}_i & \mathbf{0} \\ \mathbf{p}_i\mathbf{R}_i & \mathbf{R}_i \end{bmatrix}$, where $\mathbf{R}_i$ is the 3×3 rotation transformation matrix and $\mathbf{p}_i$ is the skew-symmetric matrix of vector $\mathbf{p}_i$ that presents the displacements of origin $O_i$ relative to origin O. Referring to polyhedral geometry and $T_d$ symmetry in Figs. 3(b) and 3(c), $\mathbf{R}_i$ and $\mathbf{p}_i$ ($i = 1, 2, \ldots, 6$) in deployable tetrahedral mechanism can be obtained and listed in Appendix A. Hence, the complete motion-screw system in the reference coordinate system can be formulated with six adjoint transformation matrices.

Furthermore, the number of links and joints involved in the deployable tetrahedral mechanism are 28 and 36, respectively. Using the Euler formula [40] for a multiloop mechanism, the number of independent loops in this mechanism can be obtained as $N_l = l - j + 1 = 36 - 28 + 1 = 9$ (with $N_l$ denoting the number of independent loops, $l$ the number of links, and $j$ the number of joints) and the associated constraint graph is sketched in Fig. 3(d). According to Kirchhoff's circulation law [41] for independent loops shown in the constraint graph, the constraint matrix of the deployable tetrahedral mechanism is organized as

$$\mathbf{M}_1 = \begin{bmatrix} S_1 & \mathbf{0}_6 & \mathbf{0}_6 & \mathbf{0}_6 & \mathbf{0}_6 & \mathbf{0}_6 \\ \mathbf{0}_6 & S_2 & \mathbf{0}_6 & \mathbf{0}_6 & \mathbf{0}_6 & \mathbf{0}_6 \\ \mathbf{0}_6 & \mathbf{0}_6 & S_3 & \mathbf{0}_6 & \mathbf{0}_6 & \mathbf{0}_6 \\ \mathbf{0}_6 & \mathbf{0}_6 & \mathbf{0}_6 & S_4 & \mathbf{0}_6 & \mathbf{0}_6 \\ \mathbf{0}_6 & \mathbf{0}_6 & \mathbf{0}_6 & \mathbf{0}_6 & S_5 & \mathbf{0}_6 \\ \mathbf{0}_6 & \mathbf{0}_6 & \mathbf{0}_6 & \mathbf{0}_6 & \mathbf{0}_6 & S_6 \\ \mathbf{0}_6 & -S_2'' & -S_3' & \mathbf{0}_6 & \mathbf{0}_6 & -S_6'' \\ -S_1'' & -S_2' & \mathbf{0}_6 & \mathbf{0}_6 & -S_5'' & \mathbf{0}_6 \\ -S_1' & \mathbf{0}_6 & -S_3'' & -S_4'' & \mathbf{0}_6 & \mathbf{0}_6 \end{bmatrix}, \quad (3)$$

where $S_i = \begin{bmatrix} S_{i1} & S_{i2} & S_{i3} & S_{i4} & S_{i5} & S_{i6} \end{bmatrix}$, $S_i' = \begin{bmatrix} S_{i1} & S_{i2} & S_{i3} & 0 & 0 & 0 \end{bmatrix}$, $S_i'' = \begin{bmatrix} 0 & 0 & 0 & S_{i4} & S_{i5} & S_{i6} \end{bmatrix}$ and $\mathbf{0}_6 = \begin{bmatrix} 0 & 0 & 0 & 0 & 0 & 0 \end{bmatrix}$ with $\mathbf{0} = \begin{bmatrix} 0 & 0 & 0 & 0 & 0 & 0 \end{bmatrix}^T$. And $S_{ij}$ ($i = 1, 2, \cdots, 9$, and $j = 1, 2, \cdots, 9$) can be obtained through Eqs. (1) and (2) together with the matrix $Ad_T$.

Mobility of this mechanism can be determined by the 54×36 constraint matrix as
$$m = n - rank(\mathbf{M}_1) = 36 - 35 = 1, \quad (4)$$
in which $m$ stands for the actual mobility of this mechanism and $n$ is the number of joints. Meanwhile, all the six involved Sarrus linkages have identical kinematic behavior to generate the synchronized radial motion of the entire mechanism, which is revealed and proved in Appendix B.

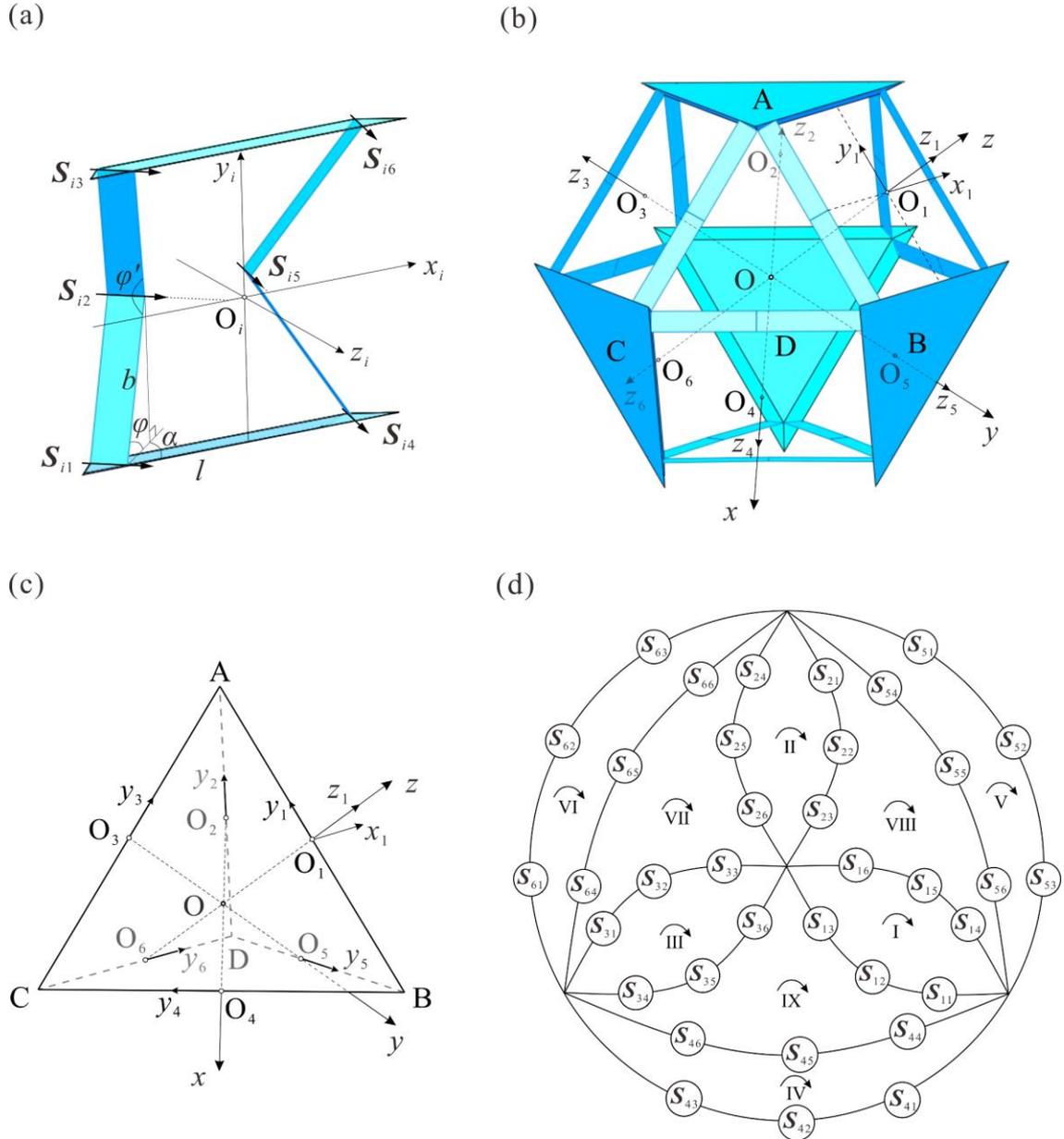

**Fig. 3.** Mobility analysis of the deployable tetrahedral mechanism. (a) Joint screws in a Sarrus linkage; coordinate systems in (b) deployed configuration of the tetrahedral mechanism and (c) its dual tetrahedron; and (d) constraint graph of this mechanism.

Based on the above derivation, we prove that the tetrahedral mechanism has mobility of one. Yet, the calculation and solution of constraint matrix are complicated due to the complexity of polyhedral geometry and a significantly large number of links and joints in the proposed polyhedral mechanism. To find a simple and effective analysis method for the mobility of deployable polyhedrons, we present the equivalent analysis strategy as follows by solving the equivalent motion screws.

Also beginning with the construction element, the constraint screw system of limb 1 in Sarrus linkage can be obtained by solving the reciprocal screws of $\mathbb{S}_{l1}$ as

$$\mathbb{S}_{l1}^r = \begin{cases} \boldsymbol{S}_{i1}^r = \begin{bmatrix} \tan\alpha & 0 & 1 & 0 & 0 & 0 \end{bmatrix}^T \\ \boldsymbol{S}_{i2}^r = \begin{bmatrix} 0 & 0 & 0 & 0 & 1 & 0 \end{bmatrix}^T \\ \boldsymbol{S}_{i3}^r = \begin{bmatrix} 0 & 0 & 0 & -\cot\alpha & 0 & 1 \end{bmatrix}^T \end{cases}, \tag{5}$$

and for limb 2, it has

$$\mathbb{S}_{l2}^r = \begin{cases} \boldsymbol{S}_{i4}^r = \begin{bmatrix} -\tan\alpha & 0 & 1 & 0 & 0 & 0 \end{bmatrix}^T \\ \boldsymbol{S}_{i5}^r = \begin{bmatrix} 0 & 0 & 0 & 0 & 1 & 0 \end{bmatrix}^T \\ \boldsymbol{S}_{i6}^r = \begin{bmatrix} 0 & 0 & 0 & \cot\alpha & 0 & 1 \end{bmatrix}^T \end{cases}. \tag{6}$$

The platform constraint-screw multiset is the combination of the above two constraint screw systems, which contains five linearly independent screws [42]. A non-unique basis for the subspace of constraint screw multiset can be selected as

$$\mathbb{S}_i^r = \begin{cases} \boldsymbol{S}_{i1}^r = \begin{bmatrix} 1 & 0 & 1 & 0 & 0 & 0 \end{bmatrix}^T \\ \boldsymbol{S}_{i2}^r = \begin{bmatrix} 0 & 0 & 0 & 0 & 1 & 0 \end{bmatrix}^T \\ \boldsymbol{S}_{i3}^r = \begin{bmatrix} 0 & 0 & 0 & -1 & 0 & 1 \end{bmatrix}^T \\ \boldsymbol{S}_{i4}^r = \begin{bmatrix} -1 & 0 & 1 & 0 & 0 & 0 \end{bmatrix}^T \\ \boldsymbol{S}_{i5}^r = \begin{bmatrix} 0 & 0 & 0 & 1 & 0 & 1 \end{bmatrix}^T \end{cases}. \tag{7}$$

By taking reciprocal screw of $\mathbb{S}_i^r$, the equivalent motion screw between two platforms in a Sarrus linkage is

$$\boldsymbol{S}_{fi} = \begin{bmatrix} 0 & 0 & 0 & 0 & 1 & 0 \end{bmatrix}^T. \tag{8}$$

which indicates the straight-line motion between two platforms along $y_i$-axis.

Hence, we can regard a Sarrus linkage as a prismatic joint, then the tetrahedral mechanism obtained in Fig. 2 can be simplified as an equivalent mechanism with six prismatic joints donated by $P_1$ to $P_6$ as indicated in Fig. 4(a). The original six motion screws in Eqs. (1) and (2) can be equivalently replaced by a single motion screw in Eq. (8), thus the equivalent topological graph is given in Fig. 4(b). Further, the simplified mobility analysis of the tetrahedral mechanism can be carried out by redrawing the constraint graph in Fig. 4(c) with equivalent motion screws $\boldsymbol{S}_{f1}$ to $\boldsymbol{S}_{f6}$, which can also be obtained in the reference coordinate system through adjoint transformation matrices. According to Fig. 4(c), the constraint matrix can be rewritten as

$$\mathbf{M}_{e1} = \begin{bmatrix} \boldsymbol{S}_{f1} & \boldsymbol{S}_{f2} & \mathbf{0} & \mathbf{0} & \boldsymbol{S}_{f5} & \mathbf{0} \\ \mathbf{0} & -\boldsymbol{S}_{f2} & \boldsymbol{S}_{f3} & \mathbf{0} & \mathbf{0} & \boldsymbol{S}_{f6} \\ -\boldsymbol{S}_{f1} & \mathbf{0} & -\boldsymbol{S}_{f3} & \boldsymbol{S}_{f4} & \mathbf{0} & \mathbf{0} \end{bmatrix}, \tag{9}$$

which has the dimension of $18 \times 6$.

Therefore, referring to this equivalent constraint matrix, the identical conclusion that the deployable tetrahedral mechanism has mobility of one can be verified as

$$m = n_e - rank(\mathbf{M}_{e1}) = 6 - 5 = 1, \tag{10}$$

in which $n_e$ is the number of equivalent prismatic joints.

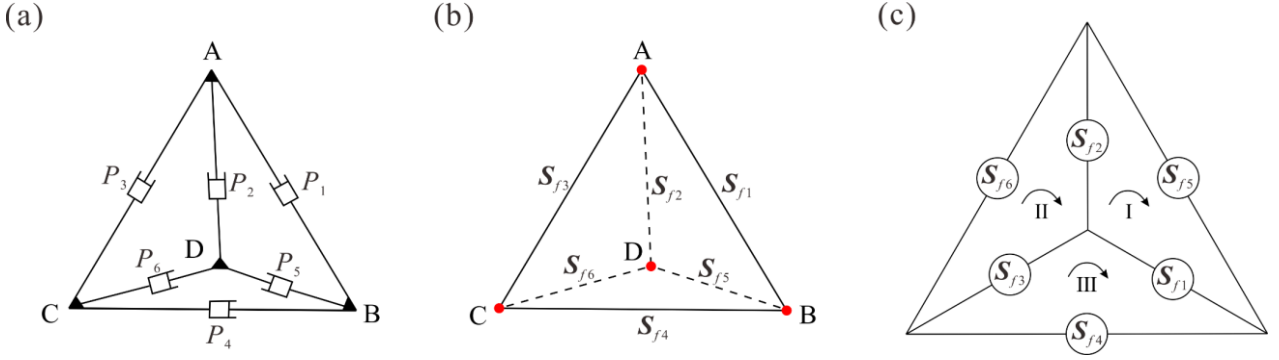

**Fig. 4.** Equivalent tetrahedral mechanism. (a) The schematic diagram of equivalent mechanism with prismatic joints, (b) its three-dimensional equivalent topological graph, and (c) the corresponding constraint graph.

### 2.3 Kinematics of the tetrahedral mechanism

Due to the one-DOF synchronized radial motion with unchanged $T_d$ symmetry of this tetrahedral mechanism, both inscribed sphere and circumscribed sphere related to four platforms are regular spheres. When $\varphi=0$ (the folding angle in a Sarrus linkage in Fig. 3(a)), referring to a tetrahedron at fully folded configuration, the inscribed sphere radius ($r$) and circumscribed sphere radius ($R$) in this mechanism are $\sqrt{6}a/12$ and $\sqrt{6}a/4$, respectively. Following deployed motion until $\varphi=90°$, i.e., a completely deployable configuration (rhombitetratetrahedron), $r$ increase to $\sqrt{6}a/3$ and $R$ becomes $a$. Thus, together with the kinematic and geometric caculation, the relationships between polyhedral geometry and kinematic angle can be derived as $r=\sqrt{6}a(3\sin\varphi+1)/12$ and $R=a\sqrt{(3\sin^2\varphi+2\sin\varphi+3)/8}$. Similarly, for the volume ($V$) of deployable tetrahedron during the continuous motion, we have $V=\sqrt{2}a^3(\sin^3\varphi+9\sin^2\varphi+9\sin\varphi+1)/12$, in which $a$ is the edge length of a regular tetrahedron. Hence, the input-output curves between the polyhedral geometry and kinematic angle are illustrated in Fig. 5.

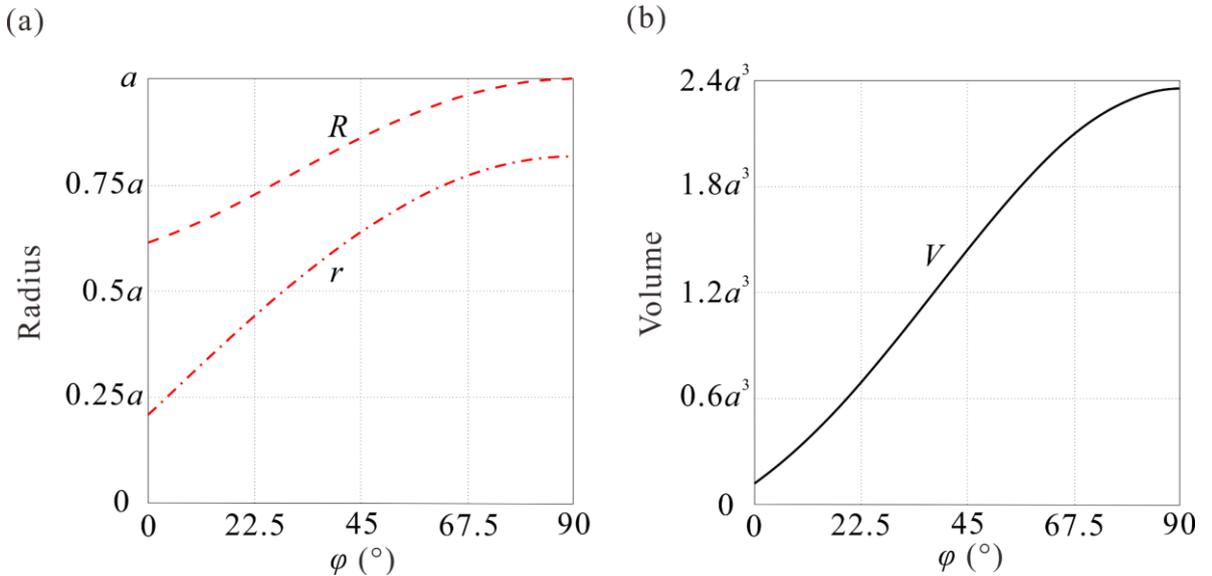

**Fig. 5.** Kinematic curves of (a) inscribed sphere radius $r$ and circumscribed sphere radius $R$ and (b) the volume $V$ vs. folding angle $\varphi$. When $\varphi$ varies from 0 to 90°, $r$ is from $\sqrt{6}a/12$ to $\sqrt{6}a/3$, $R$ is from $\sqrt{6}a/4$ to $a$, $V$ is from $\sqrt{2}a^3/12$ to $5\sqrt{2}a^3/3$.

Therefore, a novel synthesis mothed based on expansion operation and Sarrus linkages is proposed to construct the one-DOF deployable tetrahedral mechanism, its significant symmetry property is perfectly remained in each working configuration, whose corresponding prototype is fabricated, as shown in Fig. 6. The kinematic strategy of construction and mobility analysis can be readily extended to the deployable cubic and dodecahedral mechanisms with distinct symmetries in the following section.

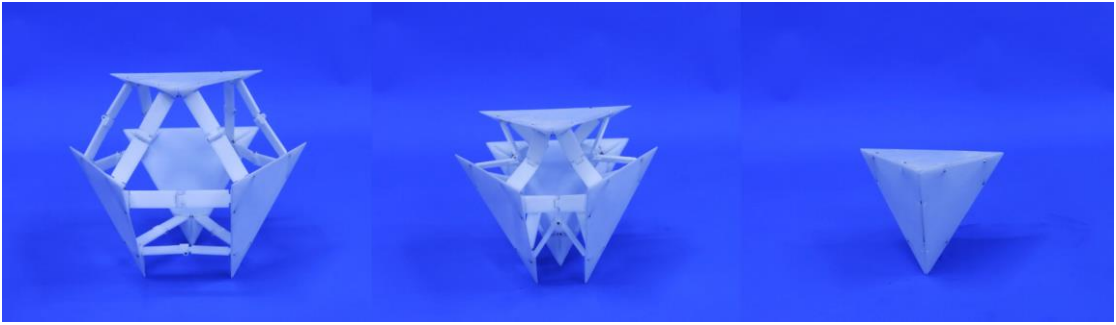

**Fig. 6.** Motion sequence of the tetrahedral mechanism.

**3 Deployable cubic and dodecahedral mechanisms**

Based on the construction of the tetrahedral mechanism in Section 2, in this section, the proposed method is extended to the design of deployable cubic and dodecahedral mechanisms in this section.

A cube with six congruous square prismoid platforms A to F and twelve edges is given in Fig. 7(a), in which the axes in a global coordinate system are perpendicular to the platforms. Following the expansion operation discussed in Section 2, Fig. 7(b) presents all separated platforms along red dash-dot perpendiculars, and a total of twelve virtual motion paths (indicated in red lines) between each pair of adjacent platforms are generated in Fig. 7(c). In this case, the angle between the bottom and side facets of each square prismoid is $\beta=45°$. As a result, twelve Sarrus linkages are needed to be involved to construct a deployable cubic mechanism, in which the geometry and kinematics of each identical Sarrus linkage are same as the one in Fig. 2(c). Meanwhile, $\gamma \in (0, 109.47°]$ should be satisfied in this case to avoid interference, and the angle between the virtual plane (Sarrus translational platform) and polyhedral platform is also $\beta=45°$. Therefore, the transformation from a rhombicuboctahedron to a cube with synchronized radial motion is obtained and shown in Fig. 7(d).

Without loss of generality, the equivalent mobility analysis method can be effectively applied to this deployable cubic mechanism involved with twelve Sarrus linkages. Similarly, by regarding a Sarrus linkage as a prismatic joint, the equivalent motion screws can be calculated based on the reference coordinate frame in Fig. 8(a) and its dual octahedron in Fig. 8(b), in which the details of adjoint transformation matrices can be found in Appendix C. Thus, the equivalent mechanism of the proposed cubic mechanism can be obtained in Fig. 8(c) with prismatic joints $P_1$ to $P_{12}$, which has a base of its dual octahedron. Inspired by Sclegel diagram [43] for polyhedral representation, i.e., a planar projection of a polyhedron, Fig. 8(d) illustrates the constraint graph of equivalent mechanism with $S_{f1}$ to $S_{f12}$, then the constraint matrix can be derived as

$$\mathbf{M}_{e2} = \begin{bmatrix} 0 & 0 & 0 & 0 & 0 & 0 & S_{f7} & 0 & 0 & 0 & S_{f11} & S_{f12} \\ 0 & 0 & S_{f3} & S_{f4} & 0 & 0 & -S_{f7} & 0 & 0 & 0 & 0 & 0 \\ 0 & 0 & 0 & 0 & 0 & S_{f6} & 0 & 0 & 0 & S_{f10} & -S_{f11} & 0 \\ 0 & 0 & 0 & 0 & 0 & 0 & 0 & S_{f8} & S_{f9} & 0 & 0 & -S_{f12} \\ S_{f1} & 0 & 0 & -S_{f4} & 0 & 0 & 0 & -S_{f8} & 0 & 0 & 0 & 0 \\ 0 & S_{f2} & -S_{f3} & 0 & 0 & -S_{f6} & 0 & 0 & 0 & 0 & 0 & 0 \\ 0 & 0 & 0 & 0 & S_{f5} & 0 & 0 & 0 & -S_{f9} & -S_{f10} & 0 & 0 \end{bmatrix}. \quad (11)$$

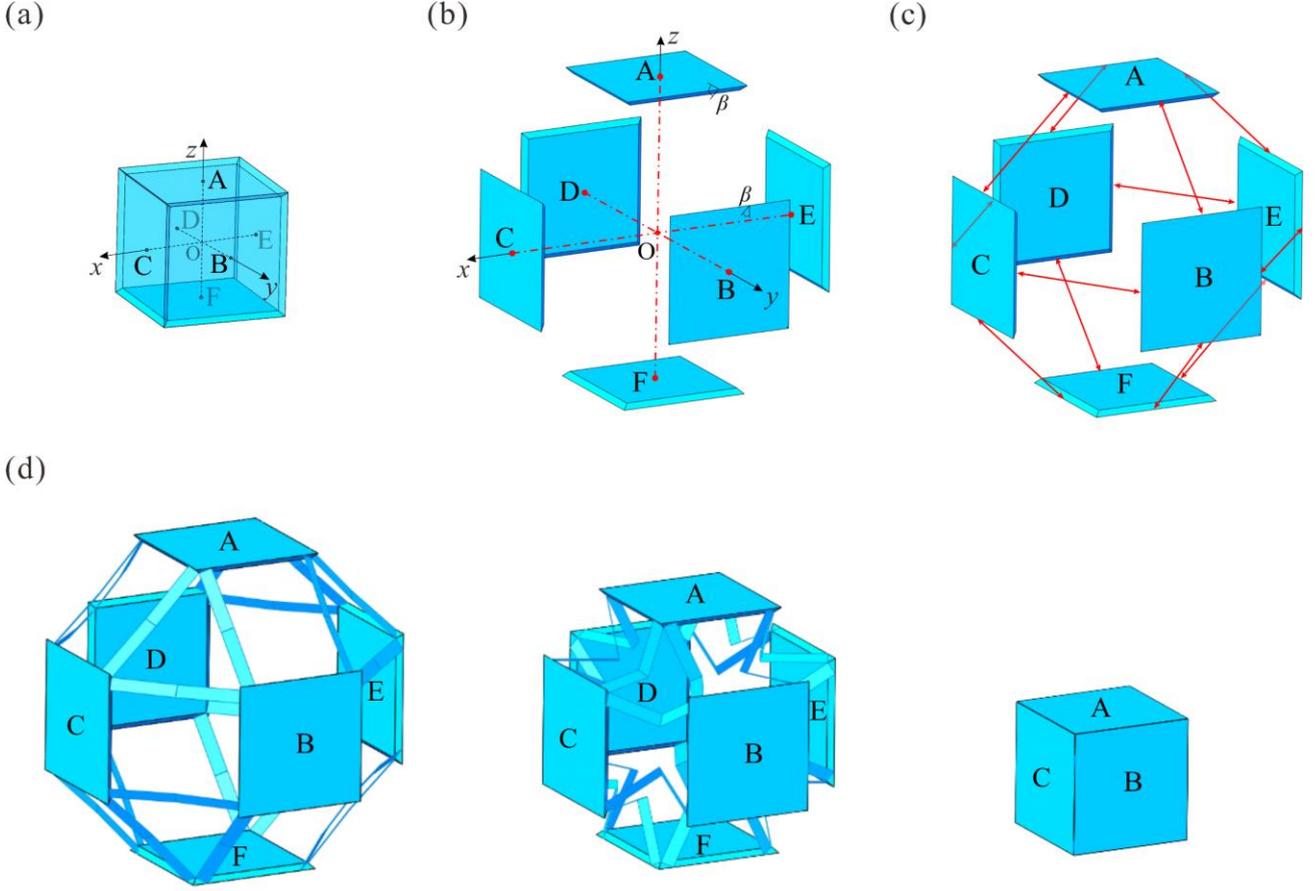

**Fig. 7.** Construction of the deployable cubic mechanism. (a) A cube and the Cartesian coordinate system; (b) the expansion of six square platforms of a cube; (c) straight-line motion paths between two adjacent platforms; (d) the motion sequence from a rhombicuboctahedron (the deployed configuration) to a cube (the folded configuration).

The rank of this $42 \times 12$ constraint matrix is 11, hence $m = n_e - rank(\mathbf{M}_{e2}) = 12 - 11 = 1$, indicating that the deployable cubic mechanism has mobility of one. To verify the equivalent analysis result, the original constraint graph and an $114 \times 72$ original constraint matrix $\mathbf{M}_2$ are given in Appendix C, from which the fact that the Sarrus-inspired deployable cubic mechanism has mobility of one is further derived.

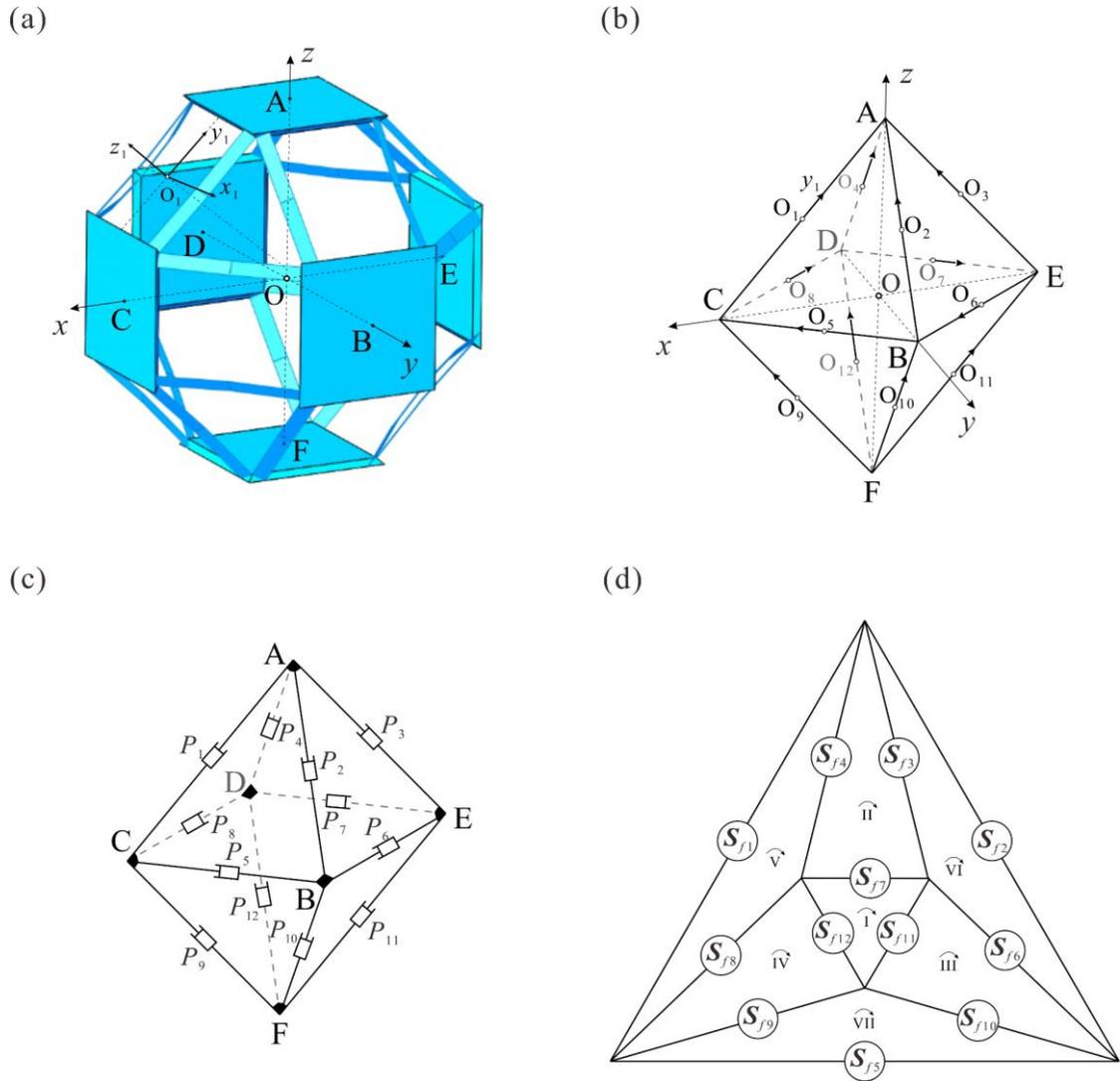

**Fig. 8.** Mobility analysis of deployable cubic mechanism. Coordinate system in (a) deployed configuration of cubic mechanism (rhombicuboctahedron) and (b) its dual octahedron ($y_i$ present the direction of straight-line motion); (c) the equivalent mechanism with twelve prismatic joints and (d) its constraint graph.

Furthermore, we also can create a deployable dodecahedral mechanism following $I_h$ symmetry with the proposed construction strategy. A dodecahedron with twelve congruous pentagonal platforms and thirty edges is given in Fig. 9(a), after the expansion operation of pentagonal prismoids with $\beta=58.28°$ ( a half of dihedral angle between two adjacent pentagonal platforms), thirty red straight-line motion paths are illustrated in Fig. 9(b). Taking the similar implantation of Sarrus linkages as given in Fig. 2(c), the deployable dodecahedron based on thirty identical Sarrus linkages is constructed in Fig. 9(c), in which $\gamma \in (0, 138.19°]$ should be considered in this mechanism Meanwhile, the radial transformation from a rhombicosidodecahedron to a dodecahedron is obtained.

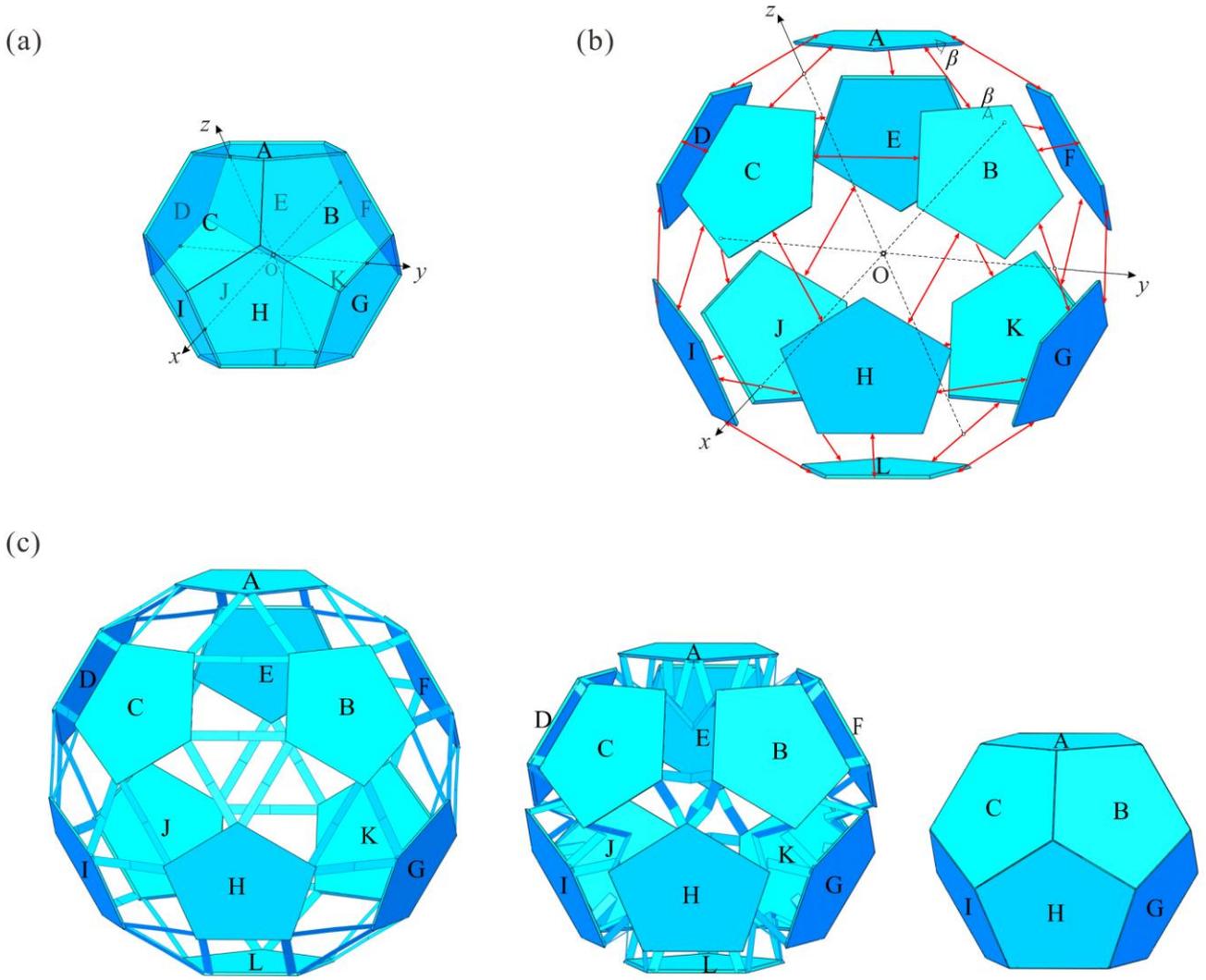

**Fig. 9.** Construction of the deployable dodecahedral mechanism. (a) A dodecahedron and the Cartesian coordinate system; (b) straight-line motion paths between two adjacent pentagonal platforms; (c) the motion sequence from a rhombicosidodecahedron (the deployed configuration) to a dodecahedron (the folded configuration).

By following the equivalent analysis approach, the reference coordinate system of the deployable dodecahedral mechanism is established in Figs. 10(a) and 10(b), thirty motion screws can be derived referring to the details listed in Appendix D. Hence, we obtain an equivalent mechanism with thirty prismatic joints on a base of dual icosahedron, see Fig. 10(c). Together with the Schlegel diagram of its dual icosahedron, the equivalent constraint graph with thirty motion screws $S_{f1}$ to $S_{f30}$ is sketched in Fig. 10(d), then an $114 \times 30$ equivalent constraint matrix $\mathbf{M}_{e3}$ can be organized as

$$\mathbf{M}_{e3} = \begin{bmatrix} \mathbf{0}_{6\times6} & \mathbf{0}_{6\times6} & \mathbf{M}_{13} & \mathbf{M}_{14} & \mathbf{M}_{15} \\ \mathbf{0}_{6\times6} & \mathbf{M}_{22} & \mathbf{M}_{23} & \mathbf{M}_{24} & \mathbf{M}_{25} \\ \mathbf{M}_{31} & \mathbf{M}_{32} & \mathbf{M}_{33} & \mathbf{0}_{7\times6} & \mathbf{0}_{7\times6} \end{bmatrix}, \qquad (12)$$

in which its submatrices are given in Appendix D.

The rank of this constraint matrix can be calculated as 29. Therefore, the mobility of the deployable dodecahedral mechanism is $m = n_e - rank(\mathbf{M}_{e3}) = 30 - 29 = 1$. Furthermore, the original constraint graph and a $294 \times 180$ original constraint matrix $\mathbf{M}_3$ of this mechanism can be found in Appendix D, as well as the same conclusion about mobility of one.

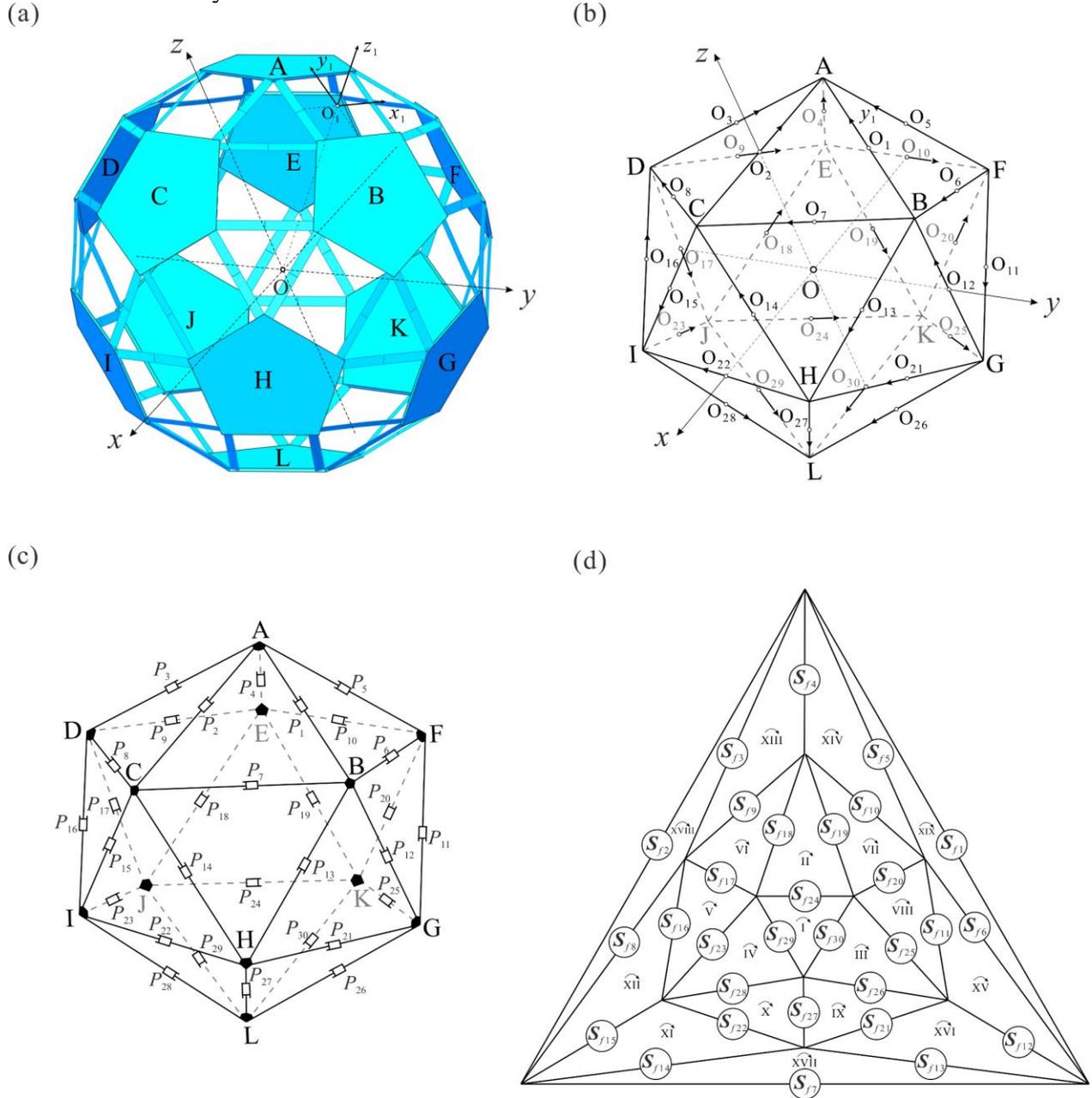

**Fig. 10.** Mobility analysis of deployable dodecahedral mechanism. Coordinate system in (a) deployed configuration of the dodecahedron (rhombicosidodecahedron) and (b) its dual icosahedron ($y_i$ present the direction of straight-line motion); (c) the equivalent mechanism with thirty prismatic joints and (d) its constraint graph.

Although only three platonic polyhedrons are investigated in this paper due to their regular transformability, such construction methods can be adapted to explore the design of other one-DOF deployable polyhedrons such as Archimedean and prismatic polyhedrons. The details of Sarrus-inspired DPMs that can be constructed with the proposed method are listed in Table 1, including the number of Sarrus linkages ($N_{\text{Sarrus}}$), links ($N_{\text{link}}$) and joints ($N_{\text{joint}}$) as well as the angle $\beta_{(M,N)}$ for radially decomposed prismoids, and the angle $\gamma_{max}$ to avoid interference. Here, a Sarrus-inspired triangular-prism mechanism is selected as an example to demonstrate the extension of proposed method, whose motion sequence is illustrated in Fig. 11. Note that, to realise such polyhedral construction, three adjacent platforms in a polyhedron should surround at a common vertex, as similar as the synthesis principle of the proposed tetrahedral, cubic and dodecahedral mechanisms.

**Table 1.** Sarrus-inspired DPMs in different polyhedral groups

| Polyhedral groups | Deployable mechanisms | $N_{Sarrus}$ | $N_{link}$ | $N_{joint}$ | $\beta_{(M,N)}$ (°) | $\gamma_{max}$ (°) |
|---|---|---|---|---|---|---|
| Platonic polyhedrons | Tetrahedron | 6 | 28 | 36 | 35.26 $_{(3,3)}$ | 70.53 |
| | Cube | 12 | 54 | 72 | 45 $_{(4,4)}$ | 109.47 |
| | Dodecahedron | 30 | 132 | 180 | 58.28 $_{(5,5)}$ | 139.18 |
| Archimedean polyhedrons | Truncated tetrahedron | 18 | 80 | 108 | 54.74 $_{(3,6)}$<br>35.26 $_{(6,6)}$ | 129.52 |
| | Truncated cube | 36 | 158 | 216 | 62.63 $_{(3,8)}$<br>45 $_{(8,8)}$ | 147.35 |
| | Truncated octahedron | 36 | 158 | 216 | 62.63 $_{(4,6)}$<br>54.74 $_{(6,6)}$ | 143.13 |
| | Truncated cuboctahedron | 72 | 314 | 432 | 72.37 $_{(4,6)}$<br>67.5 $_{(4,8)}$<br>62.63 $_{(6,8)}$ | 155.09 |
| | Truncated dodecahedron | 90 | 392 | 540 | 71.31 $_{(3,10)}$<br>58.28 $_{(10,10)}$ | 160.61 |
| | Truncated icosahedron | 90 | 392 | 540 | 69.09 $_{(6,6)}$<br>71.31 $_{(5,6)}$ | 156.72 |
| | Truncated icosidodecahedron | 180 | 782 | 1080 | 79.55 $_{(4,6)}$<br>74.14 $_{(4,10)}$<br>71.31 $_{(6,10)}$ | 164.89 |
| Prisms | $N$-prism ($N \geq 3$) | $3N$ | $13N+2$ | $18N$ | $90(N-2)/N$ $_{(4,4)}$<br>45 $_{(4,N)}$ | $2\arccos(1+\csc^2(\pi/N))^{-1/2}$ |

\* $M$ and $N$ in $\beta_{(M,N)}$ stand for two adjacent $M$-polygon and $N$-polygon.

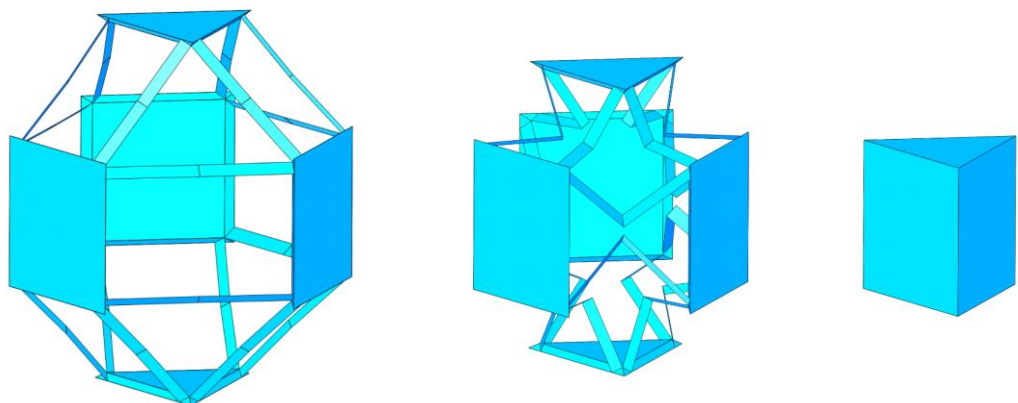

**Fig. 11.** Motion sequence of the deployable triangular-prism mechanism.

## 4 Conclusions and discussion

This paper proposed an innovative approach for constructing Sarrus-inspired deployable polyhedral mechanisms based on three Platonic polyhedrons. Through integrating the Sarrus linkages into Platonic polyhedrons after expansion operation, the deployable tetrahedral, cubic and dodecahedral mechanisms were synthesized and constructed that enable the one-DOF synchronized radial motion, in which three paired polyhedral transformations between Platonic and Archimedean polyhedrons were identified and revealed. The equivalent analysis strategy for multiloop polyhedral mechanisms was proposed by means of the equivalent prismatic joint, which brings simplicity and high-efficiency in mobility analysis.

The proposed construction and analysis strategy can be adapted to design other deployable structures in various regular and irregular polyhedral groups that could facilitate their applications in various engineering fields. Furthermore, based on the construction method in this paper, we also can realise all nine paired transformations reported in our previous work [44], where structural variations with mechanism topology isomorphism need to be carried out. Note that the proposed polyhedrons are multiloop mechanisms with a great number of overconstraints, the constraint reduction method for such multiloop mechanisms should be further investigated, in which the detailed analysis is left for another paper in the near future. Furthermore, due to the synchronized radial motion of the deployable polyhedrons, this work also paves the way to design the kinematic cell for metastructures and metamaterials, especially in $O_h$ symmetry. We expect the proposed Sarrus-inspired deployable polyhedrons and their tessellation could facilitate their applications in various engineering fields such as multifunctional metamaterials, deployable structures for architectures as well as space exploration.

**Acknowledgements**

Y. C. gratefully acknowledges financial support from the Natural Science Foundation of China (Projects No. 51825503, 52035008 and 51721003), and the Tencent Foundation (Grant XPLORER-2020-1035). X. Z. acknowledges financial support from the Natural Science Foundation of China (Project No. 52105032).

**Appendix A**

$\mathbf{R}_i$ and $\mathbf{p}_i$ ($i$ =1, 2, …, 6) of adjoint transformation matrices in deployable tetrahedral mechanism (Fig. 3) are listed as

$$\mathbf{R}_1 = \begin{bmatrix} -\sqrt{2}/2 & -\sqrt{2}/2 & 0 \\ \sqrt{2}/2 & -\sqrt{2}/2 & 0 \\ 0 & 0 & 1 \end{bmatrix}, \quad \mathbf{p}_1 = d_1 \begin{bmatrix} 0 & 0 & 1 \end{bmatrix}^{\mathrm{T}},$$

$$\mathbf{R}_2 = \begin{bmatrix} 0 & 0 & -1 \\ -\sqrt{2}/2 & -\sqrt{2}/2 & 0 \\ -\sqrt{2}/2 & \sqrt{2}/2 & 0 \end{bmatrix}, \quad \mathbf{p}_2 = d_1 \begin{bmatrix} -1 & 0 & 0 \end{bmatrix}^\mathrm{T},$$

$$\mathbf{R}_3 = \begin{bmatrix} \sqrt{2}/2 & -\sqrt{2}/2 & 0 \\ 0 & 0 & -1 \\ \sqrt{2}/2 & \sqrt{2}/2 & 0 \end{bmatrix}, \quad \mathbf{p}_3 = d_1 \begin{bmatrix} 0 & -1 & 0 \end{bmatrix}^\mathrm{T},$$

$$\mathbf{R}_4 = \begin{bmatrix} 0 & 0 & 1 \\ -\sqrt{2}/2 & -\sqrt{2}/2 & 0 \\ \sqrt{2}/2 & -\sqrt{2}/2 & 0 \end{bmatrix}, \quad \mathbf{p}_4 = d_1 \begin{bmatrix} 1 & 0 & 0 \end{bmatrix}^\mathrm{T},$$

$$\mathbf{R}_5 = \begin{bmatrix} -\sqrt{2}/2 & \sqrt{2}/2 & 0 \\ 0 & 0 & 1 \\ \sqrt{2}/2 & \sqrt{2}/2 & 0 \end{bmatrix}, \quad \mathbf{p}_5 = d_1 \begin{bmatrix} 0 & 1 & 0 \end{bmatrix}^\mathrm{T},$$

$$\mathbf{R}_6 = \begin{bmatrix} -\sqrt{2}/2 & -\sqrt{2}/2 & 0 \\ -\sqrt{2}/2 & \sqrt{2}/2 & 0 \\ 0 & 0 & -1 \end{bmatrix}, \quad \mathbf{p}_6 = d_1 \begin{bmatrix} 0 & 0 & -1 \end{bmatrix}^\mathrm{T}. \tag{A1}$$

**Appendix B**

As shown in Fig. B1, taking the Sarrus linkage between platforms A and B as an example, $\varphi_1$ and $\varphi_1'$ are two related kinematic variables (also see Fig. 3a), and $\varphi_1'=2\varphi_1$ can be easily obtained. Meanwhile, for all the six involved Sarrus linkages, $\varphi_i'=2\varphi_i$ ($i$=1 to 6). Next, among platforms A, B and C, a spatial 9$R$ linkage can be identified as an assembly of three limbs of three corresponding Sarrus linkages, in which the revolute axes $z_1$ to $z_9$ are highlighted in red. Similarly, a general kinematic solution of this spatial 9$R$ linkage is revealed in our previous work [44], including the matrices operation process based on D-H matrix method [45].

In this tetrahedral mechanism, for the spatial 9$R$ linkage among platforms A, B and C, we have the motion constraint relationships as

$$\varphi_1'=2\varphi_1, \quad \varphi_3'=2\varphi_3, \quad \varphi_4'=2\varphi_4, \tag{B1}$$

Substituting this constraint condition into kinematic solution of this 9$R$ linkage yields

$$\varphi_1=\varphi_3=\varphi_4, \quad \varphi_1'=\varphi_3'=\varphi_4'. \tag{B2}$$

Carrying out the similar calculation procedure, other constraint conditions for the rest three 9$R$ linkages are

$$\varphi_1'=2\varphi_1, \quad \varphi_2'=2\varphi_2, \quad \varphi_5'=2\varphi_5,$$
$$\varphi_2'=2\varphi_2, \quad \varphi_3'=2\varphi_3, \quad \varphi_6'=2\varphi_6,$$
$$\varphi_4'=2\varphi_4, \quad \varphi_5'=2\varphi_5, \quad \varphi_6'=2\varphi_6, \tag{B3}$$

Further, we can obtain the kinematic relationships in the entire tetrahedral mechanism as

$$\varphi_1=\varphi_2=\varphi_3=\varphi_4=\varphi_5=\varphi_6, \quad \varphi_1'=\varphi_2'=\varphi_3'=\varphi_4'=\varphi_5'=\varphi_6'. \tag{B4}$$

Therefore, all the six involved Sarrus linkages have the identical kinematic behavior that can generate the one-DOF synchronized radial motion of the entire tetrahedral mechanism.

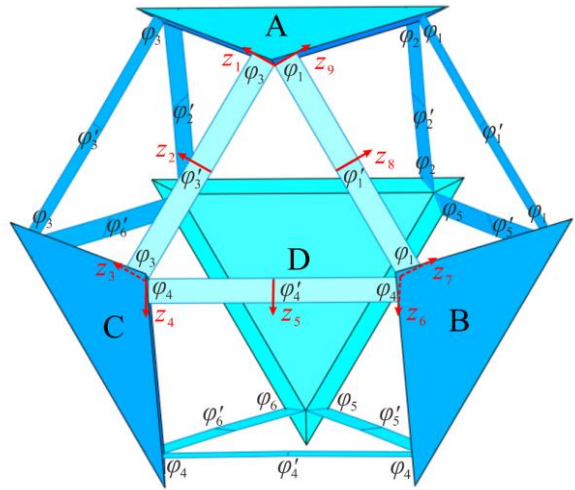

**Fig. B1.** Analysis of kinematic variables in the tetrahedral mechanism.

## Appendix C

The number of links and revolute joints in the deployable cubic mechanism are 54 and 72, respectively, in which the independent loops of this mechanism are 19. Based on the reference coordinate frame in Fig. 8 in Section 3, Fig. C1 shows the original constraint graph with 72 joint screws in the deployable cubic mechanism.

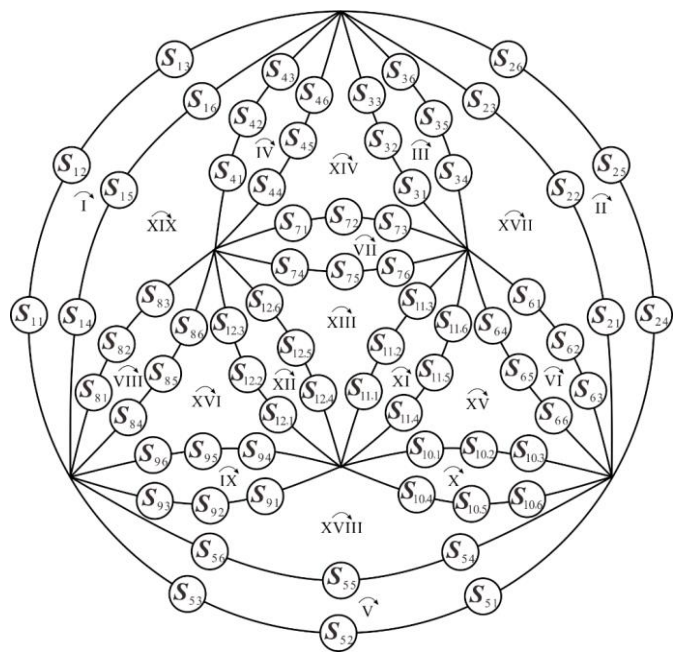

**Fig. C1.** Original constraint graph of the deployable cubic mechanism.

The details of adjoint transformation matrices are

$$\mathbf{R}_1 = \begin{bmatrix} 0 & -\sqrt{2}/2 & \sqrt{2}/2 \\ 1 & 0 & 0 \\ 0 & \sqrt{2}/2 & \sqrt{2}/2 \end{bmatrix}, \quad \mathbf{p}_1 = d_2 \begin{bmatrix} 1 & 0 & 1 \end{bmatrix}^\mathrm{T},$$

$$\mathbf{R}_2 = \begin{bmatrix} -1 & 0 & 0 \\ 0 & -\sqrt{2}/2 & \sqrt{2}/2 \\ 0 & \sqrt{2}/2 & \sqrt{2}/2 \end{bmatrix}, \quad \mathbf{p}_2 = d_2 \begin{bmatrix} 0 & 1 & 1 \end{bmatrix}^\mathrm{T},$$

$$\mathbf{R}_3 = \begin{bmatrix} 0 & \sqrt{2}/2 & -\sqrt{2}/2 \\ -1 & 0 & 0 \\ 0 & \sqrt{2}/2 & \sqrt{2}/2 \end{bmatrix}, \quad \mathbf{p}_3 = d_2 \begin{bmatrix} -1 & 0 & 1 \end{bmatrix}^\mathrm{T},$$

$$\mathbf{R}_4 = \begin{bmatrix} 1 & 0 & 0 \\ 0 & \sqrt{2}/2 & -\sqrt{2}/2 \\ 0 & \sqrt{2}/2 & \sqrt{2}/2 \end{bmatrix}, \quad \mathbf{p}_4 = d_2 \begin{bmatrix} 0 & -1 & 1 \end{bmatrix}^\mathrm{T},$$

$$\mathbf{R}_5 = \begin{bmatrix} 0 & \sqrt{2}/2 & \sqrt{2}/2 \\ 0 & -\sqrt{2}/2 & \sqrt{2}/2 \\ 1 & 0 & 0 \end{bmatrix}, \quad \mathbf{p}_5 = d_2 \begin{bmatrix} 1 & 0 & 1 \end{bmatrix}^\mathrm{T},$$

$$\mathbf{R}_6 = \begin{bmatrix} 0 & \sqrt{2}/2 & -\sqrt{2}/2 \\ 0 & \sqrt{2}/2 & \sqrt{2}/2 \\ 1 & 0 & 0 \end{bmatrix}, \quad \mathbf{p}_6 = d_2 \begin{bmatrix} -1 & 1 & 0 \end{bmatrix}^\mathrm{T},$$

$$\mathbf{R}_7 = \begin{bmatrix} 0 & -\sqrt{2}/2 & -\sqrt{2}/2 \\ 0 & \sqrt{2}/2 & -\sqrt{2}/2 \\ 1 & 0 & 0 \end{bmatrix}, \quad \mathbf{p}_7 = d_2 \begin{bmatrix} -1 & -1 & 0 \end{bmatrix}^\mathrm{T},$$

$$\mathbf{R}_8 = \begin{bmatrix} 0 & -\sqrt{2}/2 & \sqrt{2}/2 \\ 0 & -\sqrt{2}/2 & -\sqrt{2}/2 \\ 1 & 0 & 0 \end{bmatrix}, \quad \mathbf{p}_8 = d_2 \begin{bmatrix} 1 & -1 & 0 \end{bmatrix}^\mathrm{T},$$

$$\mathbf{R}_9 = \begin{bmatrix} 0 & \sqrt{2}/2 & \sqrt{2}/2 \\ 1 & 0 & 0 \\ 0 & \sqrt{2}/2 & -\sqrt{2}/2 \end{bmatrix}, \quad \mathbf{p}_9 = d_2 \begin{bmatrix} 1 & 0 & -1 \end{bmatrix}^\mathrm{T},$$

$$\mathbf{R}_{10} = \begin{bmatrix} -1 & 0 & 0 \\ 0 & \sqrt{2}/2 & \sqrt{2}/2 \\ 0 & \sqrt{2}/2 & -\sqrt{2}/2 \end{bmatrix}, \quad \mathbf{p}_{10} = d_2 \begin{bmatrix} 0 & 1 & -1 \end{bmatrix}^\mathrm{T},$$

$$\mathbf{R}_{11} = \begin{bmatrix} 0 & -\sqrt{2}/2 & -\sqrt{2}/2 \\ -1 & 0 & 0 \\ 0 & \sqrt{2}/2 & -\sqrt{2}/2 \end{bmatrix}, \quad \mathbf{p}_{11} = d_2 \begin{bmatrix} -1 & 0 & -1 \end{bmatrix}^\mathrm{T},$$

$$\mathbf{R}_{12} = \begin{bmatrix} 1 & 0 & 0 \\ 0 & -\sqrt{2}/2 & -\sqrt{2}/2 \\ 0 & \sqrt{2}/2 & -\sqrt{2}/2 \end{bmatrix}, \quad \mathbf{p}_{12} = d_2 \begin{bmatrix} 0 & -1 & -1 \end{bmatrix}^\mathrm{T}, \quad \text{(C1)}$$

and the $114 \times 72$ original constraint matrix $\mathbf{M}_2$ can be expressed as

$$\mathbf{M}_2 = \begin{bmatrix} \mathbf{M}_{11} & \mathbf{0}_{6\times 36} \\ \mathbf{0}_{6\times 36} & \mathbf{M}_{22} \\ \mathbf{M}_{31} & \mathbf{M}_{32} \end{bmatrix}, \tag{C2}$$

where $\mathbf{0}_{6\times 36} = \begin{bmatrix} \mathbf{0}_6 & \mathbf{0}_6 & \mathbf{0}_6 & \mathbf{0}_6 & \mathbf{0}_6 & \mathbf{0}_6 \\ \mathbf{0}_6 & \mathbf{0}_6 & \mathbf{0}_6 & \mathbf{0}_6 & \mathbf{0}_6 & \mathbf{0}_6 \\ \mathbf{0}_6 & \mathbf{0}_6 & \mathbf{0}_6 & \mathbf{0}_6 & \mathbf{0}_6 & \mathbf{0}_6 \\ \mathbf{0}_6 & \mathbf{0}_6 & \mathbf{0}_6 & \mathbf{0}_6 & \mathbf{0}_6 & \mathbf{0}_6 \\ \mathbf{0}_6 & \mathbf{0}_6 & \mathbf{0}_6 & \mathbf{0}_6 & \mathbf{0}_6 & \mathbf{0}_6 \\ \mathbf{0}_6 & \mathbf{0}_6 & \mathbf{0}_6 & \mathbf{0}_6 & \mathbf{0}_6 & \mathbf{0}_6 \end{bmatrix}$, $\mathbf{M}_{11} = \begin{bmatrix} S_1 & \mathbf{0}_6 & \mathbf{0}_6 & \mathbf{0}_6 & \mathbf{0}_6 & \mathbf{0}_6 \\ \mathbf{0}_6 & S_2 & \mathbf{0}_6 & \mathbf{0}_6 & \mathbf{0}_6 & \mathbf{0}_6 \\ \mathbf{0}_6 & \mathbf{0}_6 & S_3 & \mathbf{0}_6 & \mathbf{0}_6 & \mathbf{0}_6 \\ \mathbf{0}_6 & \mathbf{0}_6 & \mathbf{0}_6 & S_4 & \mathbf{0}_6 & \mathbf{0}_6 \\ \mathbf{0}_6 & \mathbf{0}_6 & \mathbf{0}_6 & \mathbf{0}_6 & S_5 & \mathbf{0}_6 \\ \mathbf{0}_6 & \mathbf{0}_6 & \mathbf{0}_6 & \mathbf{0}_6 & \mathbf{0}_6 & S_6 \end{bmatrix}$,

$\mathbf{M}_{22} = \begin{bmatrix} S_7 & \mathbf{0}_6 & \mathbf{0}_6 & \mathbf{0}_6 & \mathbf{0}_6 & \mathbf{0}_6 \\ \mathbf{0}_6 & S_8 & \mathbf{0}_6 & \mathbf{0}_6 & \mathbf{0}_6 & \mathbf{0}_6 \\ \mathbf{0}_6 & \mathbf{0}_6 & S_9 & \mathbf{0}_6 & \mathbf{0}_6 & \mathbf{0}_6 \\ \mathbf{0}_6 & \mathbf{0}_6 & \mathbf{0}_6 & S_{10} & \mathbf{0}_6 & \mathbf{0}_6 \\ \mathbf{0}_6 & \mathbf{0}_6 & \mathbf{0}_6 & \mathbf{0}_6 & S_{11} & \mathbf{0}_6 \\ \mathbf{0}_6 & \mathbf{0}_6 & \mathbf{0}_6 & \mathbf{0}_6 & \mathbf{0}_6 & S_{12} \end{bmatrix}$, $\mathbf{M}_{31} = \begin{bmatrix} \mathbf{0}_6 & \mathbf{0}_6 & \mathbf{0}_6 & \mathbf{0}_6 & \mathbf{0}_6 & \mathbf{0}_6 \\ \mathbf{0}_6 & \mathbf{0}_6 & \mathbf{0}_6 & \mathbf{0}_6 & \mathbf{0}_6 & \mathbf{0}_6 \\ \mathbf{0}_6 & \mathbf{0}_6 & -S_3' & -S_4'' & \mathbf{0}_6 & \mathbf{0}_6 \\ \mathbf{0}_6 & \mathbf{0}_6 & \mathbf{0}_6 & \mathbf{0}_6 & \mathbf{0}_6 & -S_6'' \\ \mathbf{0}_6 & -S_2' & -S_3'' & \mathbf{0}_6 & \mathbf{0}_6 & -S_6' \\ \mathbf{0}_6 & \mathbf{0}_6 & \mathbf{0}_6 & \mathbf{0}_6 & -S_5'' & \mathbf{0}_6 \\ -S_1'' & \mathbf{0}_6 & \mathbf{0}_6 & -S_4' & \mathbf{0}_6 & \mathbf{0}_6 \end{bmatrix}$,

$\mathbf{M}_{32} = \begin{bmatrix} -S_7'' & \mathbf{0}_6 & \mathbf{0}_6 & \mathbf{0}_6 & -S_{11}' & -S_{12}'' \\ \mathbf{0}_6 & -S_8'' & -S_9'' & \mathbf{0}_6 & \mathbf{0}_6 & -S_{12}' \\ -S_7' & \mathbf{0}_6 & \mathbf{0}_6 & \mathbf{0}_6 & \mathbf{0}_6 & \mathbf{0}_6 \\ \mathbf{0}_6 & \mathbf{0}_6 & \mathbf{0}_6 & -S_{10}' & -S_{11}'' & \mathbf{0}_6 \\ \mathbf{0}_6 & \mathbf{0}_6 & \mathbf{0}_6 & \mathbf{0}_6 & \mathbf{0}_6 & \mathbf{0}_6 \\ \mathbf{0}_6 & \mathbf{0}_6 & -S_9' & -S_{10}'' & \mathbf{0}_6 & \mathbf{0}_6 \\ \mathbf{0}_6 & -S_8' & \mathbf{0}_6 & \mathbf{0}_6 & \mathbf{0}_6 & \mathbf{0}_6 \end{bmatrix}$.

Thus, the rank of the original constraint matrix $\mathbf{M}_2$ is 71, which indicates the mobility of the deployable cubic mechanism as $m = n - rank(\mathbf{M}_2) = 72 - 71 = 1$.

**Appendix D**

Firstly, the details of submatrices of $\mathbf{M}_{e3}$ (Eq. (12) in Section 3) can be listed as

$$\mathbf{M}_{13} = \begin{bmatrix} 0 & 0 & 0 & 0 & 0 & 0 \\ 0 & 0 & 0 & 0 & 0 & S_{f18} \\ 0 & 0 & 0 & 0 & 0 & 0 \\ 0 & 0 & 0 & 0 & 0 & 0 \\ 0 & 0 & 0 & S_{f16} & S_{f17} & 0 \\ 0 & S_{f14} & S_{f15} & 0 & 0 & 0 \end{bmatrix},$$

$$\mathbf{M}_{14} = \begin{bmatrix} 0 & 0 & 0 & 0 & 0 & S_{f24} \\ S_{f19} & 0 & 0 & 0 & 0 & -S_{f24} \\ 0 & 0 & 0 & 0 & 0 & 0 \\ 0 & 0 & 0 & 0 & S_{f23} & 0 \\ 0 & 0 & 0 & 0 & -S_{f23} & 0 \\ 0 & 0 & 0 & S_{f22} & 0 & 0 \end{bmatrix},$$

$$\mathbf{M}_{15} = \begin{bmatrix} 0 & 0 & 0 & 0 & S_{f29} & S_{f30} \\ 0 & 0 & 0 & 0 & 0 & 0 \\ S_{f25} & S_{f26} & 0 & 0 & 0 & -S_{f30} \\ 0 & 0 & 0 & S_{f28} & -S_{f29} & 0 \\ 0 & 0 & 0 & 0 & 0 & 0 \\ 0 & 0 & 0 & 0 & 0 & 0 \end{bmatrix},$$

$$\mathbf{M}_{22} = \begin{bmatrix} 0 & 0 & 0 & S_{f10} & 0 & 0 \\ 0 & 0 & 0 & 0 & S_{f11} & 0 \\ 0 & 0 & 0 & 0 & 0 & 0 \\ 0 & 0 & 0 & 0 & 0 & 0 \\ 0 & 0 & S_{f9} & 0 & 0 & 0 \\ 0 & 0 & 0 & 0 & 0 & S_{f12} \end{bmatrix},$$

$$\mathbf{M}_{23} = \begin{bmatrix} 0 & 0 & 0 & 0 & 0 & 0 \\ 0 & 0 & 0 & 0 & 0 & 0 \\ 0 & 0 & 0 & 0 & 0 & 0 \\ 0 & 0 & 0 & 0 & 0 & 0 \\ 0 & 0 & 0 & -S_{f17} & -S_{f18} & 0 \\ S_{f13} & 0 & 0 & 0 & 0 & 0 \end{bmatrix},$$

$$\mathbf{M}_{24} = \begin{bmatrix} -S_{f19} & S_{f20} & 0 & 0 & 0 & 0 \\ 0 & -S_{f20} & 0 & 0 & 0 & 0 \\ 0 & 0 & S_{f21} & 0 & 0 & 0 \\ 0 & 0 & 0 & -S_{f22} & 0 & 0 \\ 0 & 0 & 0 & 0 & 0 & 0 \\ 0 & 0 & -S_{f21} & 0 & 0 & 0 \end{bmatrix},$$

$$\mathbf{M}_{25} = \begin{bmatrix} 0 & 0 & 0 & 0 & 0 & 0 \\ -S_{f25} & 0 & 0 & 0 & 0 & 0 \\ 0 & -S_{f26} & S_{f27} & 0 & 0 & 0 \\ 0 & 0 & -S_{f27} & -S_{f28} & 0 & 0 \\ 0 & 0 & 0 & 0 & 0 & 0 \\ 0 & 0 & 0 & 0 & 0 & 0 \end{bmatrix},$$

$$\mathbf{M}_{31} = \begin{bmatrix} 0 & 0 & S_{f3} & S_{f4} & 0 & 0 \\ 0 & 0 & 0 & -S_{f4} & -S_{f5} & 0 \\ 0 & 0 & 0 & 0 & 0 & S_{f6} \\ 0 & 0 & 0 & 0 & 0 & 0 \\ 0 & 0 & 0 & 0 & 0 & 0 \\ 0 & S_{f2} & -S_{f3} & 0 & 0 & 0 \\ S_{f1} & 0 & 0 & 0 & -S_{f5} & -S_{f6} \end{bmatrix},$$

$$\mathbf{M}_{32} = \begin{bmatrix} 0 & 0 & -S_{f9} & 0 & 0 & 0 \\ 0 & 0 & 0 & -S_{f10} & 0 & 0 \\ 0 & 0 & 0 & 0 & -S_{f11} & -S_{f12} \\ 0 & S_{f8} & 0 & 0 & 0 & 0 \\ S_{f7} & 0 & 0 & 0 & 0 & 0 \\ 0 & -S_{f8} & 0 & 0 & 0 & 0 \\ 0 & 0 & 0 & 0 & 0 & 0 \end{bmatrix},$$

$$\mathbf{M}_{33} = \begin{bmatrix} 0 & 0 & 0 & 0 & 0 & 0 \\ 0 & 0 & 0 & 0 & 0 & 0 \\ 0 & 0 & 0 & 0 & 0 & 0 \\ 0 & 0 & -S_{f15} & -S_{f16} & 0 & 0 \\ -S_{f13} & -S_{f14} & 0 & 0 & 0 & 0 \\ 0 & 0 & 0 & 0 & 0 & 0 \\ 0 & 0 & 0 & 0 & 0 & 0 \end{bmatrix}.$$

Next, the deployable dodecahedral mechanism consists of 132 links and 180 revolute joints, and the independent loops of this mechanism are 49. Referring to the reference coordinate frame in Fig. 10 in Section 3, the original constraint graph with 180 joint screws is shown in Fig. D1.

The details of adjoint transformation matrices in this mechanism are

$$\mathbf{R}_1 = \begin{bmatrix} -(\sqrt{5}+1)/4 & -1/2 & -(\sqrt{5}-1)/4 \\ (\sqrt{5}-1)/4 & -(\sqrt{5}+1)/4 & 1/2 \\ -1/2 & (\sqrt{5}-1)/4 & (\sqrt{5}+1)/4 \end{bmatrix}, \quad \mathbf{p}_1 = d_3 \begin{bmatrix} -(\sqrt{5}-1)/4 \\ 1/2 \\ (\sqrt{5}+1)/4 \end{bmatrix},$$

$$\mathbf{R}_2 = \begin{bmatrix} 0 & -1 & 0 \\ 1 & 0 & 0 \\ 0 & 0 & 1 \end{bmatrix}, \quad \mathbf{p}_2 = d_3 \begin{bmatrix} 0 \\ 0 \\ 1 \end{bmatrix},$$

$$\mathbf{R}_3 = \begin{bmatrix} (\sqrt{5}+1)/4 & -1/2 & -(\sqrt{5}-1)/4 \\ (\sqrt{5}-1)/4 & (\sqrt{5}+1)/4 & -1/2 \\ 1/2 & (\sqrt{5}-1)/4 & (\sqrt{5}+1)/4 \end{bmatrix}, \quad \mathbf{p}_3 = d_3 \begin{bmatrix} -(\sqrt{5}-1)/4 \\ -1/2 \\ (\sqrt{5}+1)/4 \end{bmatrix},$$

$$\mathbf{R}_4 = \begin{bmatrix} 1/2 & (\sqrt{5}-1)/4 & -(\sqrt{5}+1)/4 \\ -(\sqrt{5}+1)/4 & 1/2 & -(\sqrt{5}-1)/4 \\ (\sqrt{5}-1)/4 & (\sqrt{5}+1)/4 & 1/2 \end{bmatrix}, \quad \mathbf{p}_4 = d_3 \begin{bmatrix} -(\sqrt{5}+1)/4 \\ -(\sqrt{5}-1)/4 \\ 1/2 \end{bmatrix}$$

$$\mathbf{R}_5 = \begin{bmatrix} -1/2 & (\sqrt{5}-1)/4 & -(\sqrt{5}+1)/4 \\ -(\sqrt{5}+1)/4 & -1/2 & (\sqrt{5}-1)/4 \\ -(\sqrt{5}-1)/4 & (\sqrt{5}+1)/4 & 1/2 \end{bmatrix}, \quad \mathbf{p}_5 = d_3 \begin{bmatrix} -(\sqrt{5}+1)/4 \\ (\sqrt{5}-1)/4 \\ 1/2 \end{bmatrix},$$

$$\mathbf{R}_6 = \begin{bmatrix} -(\sqrt{5}-1)/4 & (\sqrt{5}+1)/4 & -1/2 \\ -1/2 & (\sqrt{5}-1)/4 & (\sqrt{5}+1)/4 \\ (\sqrt{5}+1)/4 & 1/2 & (\sqrt{5}-1)/4 \end{bmatrix}, \quad \mathbf{p}_6 = d_3 \begin{bmatrix} -1/2 \\ (\sqrt{5}+1)/4 \\ (\sqrt{5}-1)/4 \end{bmatrix},$$

$$\mathbf{R}_7 = \begin{bmatrix} -(\sqrt{5}+1)/4 & 1/2 & (\sqrt{5}-1)/4 \\ -(\sqrt{5}-1)/4 & -(\sqrt{5}+1)/4 & 1/2 \\ 1/2 & (\sqrt{5}-1)/4 & (\sqrt{5}+1)/4 \end{bmatrix}, \quad \mathbf{p}_7 = d_3 \begin{bmatrix} (\sqrt{5}-1)/4 \\ 1/2 \\ (\sqrt{5}+1)/4 \end{bmatrix},$$

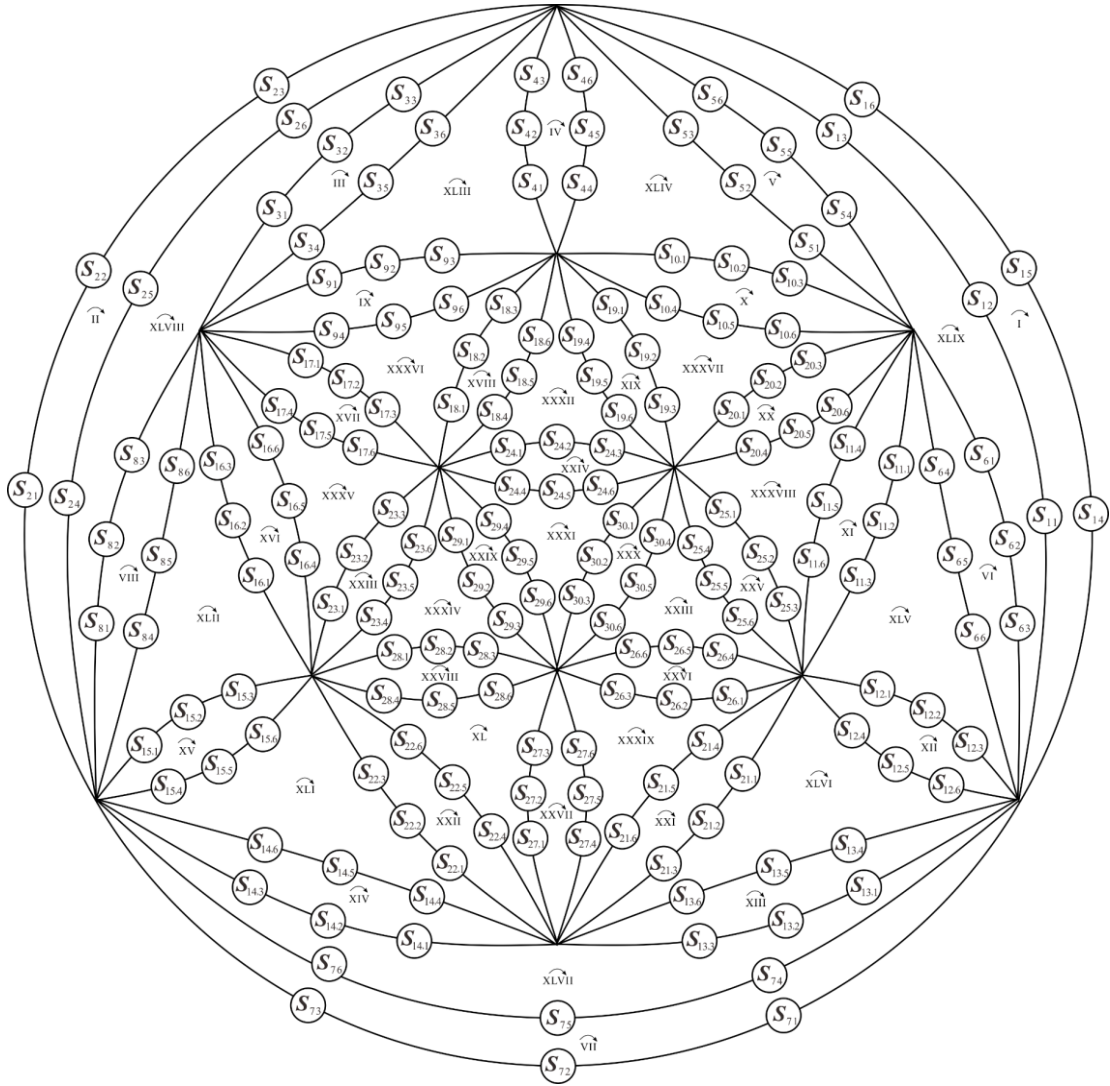

**Fig. D1.** Original constraint graph of the deployable dodecahedral mechanism.

$$\mathbf{R}_8 = \begin{bmatrix} -(\sqrt{5}+1)/4 & -1/2 & (\sqrt{5}-1)/4 \\ (\sqrt{5}-1)/4 & -(\sqrt{5}+1)/4 & -1/2 \\ 1/2 & -(\sqrt{5}-1)/4 & (\sqrt{5}+1)/4 \end{bmatrix}, \quad \mathbf{p}_8 = d_3 \begin{bmatrix} (\sqrt{5}-1)/4 \\ -1/2 \\ (\sqrt{5}+1)/4 \end{bmatrix},$$

$$\mathbf{R}_9 = \begin{bmatrix} -(\sqrt{5}-1)/4 & -(\sqrt{5}+1)/4 & -1/2 \\ 1/2 & (\sqrt{5}-1)/4 & -(\sqrt{5}+1)/4 \\ (\sqrt{5}+1)/4 & -1/2 & (\sqrt{5}-1)/4 \end{bmatrix}, \quad \mathbf{p}_9 = d_3 \begin{bmatrix} -1/2 \\ -(\sqrt{5}+1)/4 \\ (\sqrt{5}-1)/4 \end{bmatrix},$$

$$\mathbf{R}_{10} = \begin{bmatrix} 0 & 0 & -1 \\ 0 & 1 & 0 \\ 1 & 0 & 0 \end{bmatrix}, \quad \mathbf{p}_{10} = d_3 \begin{bmatrix} -1 \\ 0 \\ 0 \end{bmatrix},$$

$$\mathbf{R}_{11} = \begin{bmatrix} (\sqrt{5}-1)/4 & (\sqrt{5}+1)/4 & -1/2 \\ 1/2 & (\sqrt{5}-1)/4 & (\sqrt{5}+1)/4 \\ (\sqrt{5}+1)/4 & -1/2 & -(\sqrt{5}-1)/4 \end{bmatrix}, \quad \mathbf{p}_{11} = d_3 \begin{bmatrix} -1/2 \\ (\sqrt{5}+1)/4 \\ -(\sqrt{5}-1)/4 \end{bmatrix},$$

$$\mathbf{R}_{12} = \begin{bmatrix} -1 & 0 & 0 \\ 0 & 0 & 1 \\ 0 & 1 & 0 \end{bmatrix}, \quad \mathbf{p}_{12} = d_3 \begin{bmatrix} 0 \\ 1 \\ 0 \end{bmatrix},$$

$$\mathbf{R}_{13} = \begin{bmatrix} (\sqrt{5}-1)/4 & (\sqrt{5}+1)/4 & 1/2 \\ -1/2 & -(\sqrt{5}-1)/4 & (\sqrt{5}+1)/4 \\ (\sqrt{5}+1)/4 & -1/2 & (\sqrt{5}-1)/4 \end{bmatrix}, \quad \mathbf{p}_{13} = d_3 \begin{bmatrix} 1/2 \\ (\sqrt{5}+1)/4 \\ (\sqrt{5}-1)/4 \end{bmatrix},$$

$$\mathbf{R}_{14} = \begin{bmatrix} -1/2 & -(\sqrt{5}-1)/4 & (\sqrt{5}+1)/4 \\ (\sqrt{5}+1)/4 & -1/2 & (\sqrt{5}-1)/4 \\ (\sqrt{5}-1)/4 & (\sqrt{5}+1)/4 & 1/2 \end{bmatrix}, \quad \mathbf{p}_{14} = d_3 \begin{bmatrix} (\sqrt{5}+1)/4 \\ (\sqrt{5}-1)/4 \\ 1/2 \end{bmatrix},$$

$$\mathbf{R}_{15} = \begin{bmatrix} -1/2 & (\sqrt{5}-1)/4 & (\sqrt{5}+1)/4 \\ -(\sqrt{5}+1)/4 & -1/2 & -(\sqrt{5}-1)/4 \\ (\sqrt{5}-1)/4 & -(\sqrt{5}+1)/4 & 1/2 \end{bmatrix}, \quad \mathbf{p}_{15} = d_3 \begin{bmatrix} (\sqrt{5}+1)/4 \\ -(\sqrt{5}-1)/4 \\ 1/2 \end{bmatrix},$$

$$\mathbf{R}_{16} = \begin{bmatrix} (\sqrt{5}-1)/4 & -(\sqrt{5}+1)/4 & 1/2 \\ 1/2 & -(\sqrt{5}-1)/4 & -(\sqrt{5}+1)/4 \\ (\sqrt{5}+1)/4 & -1/2 & (\sqrt{5}-1)/4 \end{bmatrix}, \quad \mathbf{p}_{16} = d_3 \begin{bmatrix} 1/2 \\ -(\sqrt{5}+1)/4 \\ (\sqrt{5}-1)/4 \end{bmatrix},$$

$$\mathbf{R}_{17} = \begin{bmatrix} -1 & 0 & 0 \\ 0 & 0 & -1 \\ 0 & -1 & 0 \end{bmatrix}, \quad \mathbf{p}_{17} = d_3 \begin{bmatrix} 0 \\ -1 \\ 0 \end{bmatrix},$$

$$\mathbf{R}_{18} = \begin{bmatrix} (\sqrt{5}-1)/4 & -(\sqrt{5}+1)/4 & -1/2 \\ -1/2 & (\sqrt{5}-1)/4 & -(\sqrt{5}+1)/4 \\ (\sqrt{5}+1)/4 & 1/2 & -(\sqrt{5}-1)/4 \end{bmatrix}, \quad \mathbf{p}_{18} = d_3 \begin{bmatrix} -1/2 \\ -(\sqrt{5}+1)/4 \\ -(\sqrt{5}-1)/4 \end{bmatrix},$$

$$\mathbf{R}_{19} = \begin{bmatrix} -1/2 & (\sqrt{5}-1)/4 & -(\sqrt{5}+1)/4 \\ (\sqrt{5}+1)/4 & 1/2 & -(\sqrt{5}-1)/4 \\ (\sqrt{5}-1)/4 & -(\sqrt{5}+1)/4 & -1/2 \end{bmatrix}, \quad \boldsymbol{p}_{19} = d_3 \begin{bmatrix} -(\sqrt{5}+1)/4 \\ -(\sqrt{5}-1)/4 \\ -1/2 \end{bmatrix},$$

$$\mathbf{R}_{20} = \begin{bmatrix} -1/2 & -(\sqrt{5}-1)/4 & -(\sqrt{5}+1)/4 \\ -(\sqrt{5}+1)/4 & 1/2 & (\sqrt{5}-1)/4 \\ (\sqrt{5}-1)/4 & (\sqrt{5}+1)/4 & -1/2 \end{bmatrix}, \quad \boldsymbol{p}_{20} = d_3 \begin{bmatrix} -(\sqrt{5}+1)/4 \\ (\sqrt{5}-1)/4 \\ -1/2 \end{bmatrix},$$

$$\mathbf{R}_{21} = \begin{bmatrix} -(\sqrt{5}-1)/4 & (\sqrt{5}+1)/4 & 1/2 \\ 1/2 & -(\sqrt{5}-1)/4 & (\sqrt{5}+1)/4 \\ (\sqrt{5}+1)/4 & 1/2 & -(\sqrt{5}-1)/4 \end{bmatrix}, \quad \boldsymbol{p}_{21} = d_3 \begin{bmatrix} 1/2 \\ (\sqrt{5}+1)/4 \\ -(\sqrt{5}-1)/4 \end{bmatrix},$$

$$\mathbf{R}_{22} = \begin{bmatrix} 0 & 0 & 1 \\ 0 & -1 & 0 \\ 1 & 0 & 0 \end{bmatrix}, \quad \boldsymbol{p}_{22} = d_3 \begin{bmatrix} 1 \\ 0 \\ 0 \end{bmatrix},$$

$$\mathbf{R}_{23} = \begin{bmatrix} -(\sqrt{5}-1)/4 & -(\sqrt{5}+1)/4 & 1/2 \\ -1/2 & -(\sqrt{5}-1)/4 & -(\sqrt{5}+1)/4 \\ (\sqrt{5}+1)/4 & -1/2 & -(\sqrt{5}-1)/4 \end{bmatrix}, \quad \boldsymbol{p}_{23} = d_3 \begin{bmatrix} 1/2 \\ -(\sqrt{5}+1)/4 \\ -(\sqrt{5}-1)/4 \end{bmatrix},$$

$$\mathbf{R}_{24} = \begin{bmatrix} -(\sqrt{5}+1)/4 & -1/2 & -(\sqrt{5}-1)/4 \\ -(\sqrt{5}-1)/4 & (\sqrt{5}+1)/4 & -1/2 \\ 1/2 & -(\sqrt{5}-1)/4 & -(\sqrt{5}+1)/4 \end{bmatrix}, \quad \boldsymbol{p}_{24} = d_3 \begin{bmatrix} -(\sqrt{5}-1)/4 \\ -1/2 \\ -(\sqrt{5}+1)/4 \end{bmatrix},$$

$$\mathbf{R}_{25} = \begin{bmatrix} -(\sqrt{5}+1)/4 & 1/2 & -(\sqrt{5}-1)/4 \\ (\sqrt{5}-1)/4 & (\sqrt{5}+1)/4 & 1/2 \\ 1/2 & (\sqrt{5}-1)/4 & -(\sqrt{5}+1)/4 \end{bmatrix}, \quad \boldsymbol{p}_{25} = d_3 \begin{bmatrix} -(\sqrt{5}-1)/4 \\ 1/2 \\ -(\sqrt{5}+1)/4 \end{bmatrix},$$

$$\mathbf{R}_{26} = \begin{bmatrix} (\sqrt{5}+1)/4 & 1/2 & (\sqrt{5}-1)/4 \\ (\sqrt{5}-1)/4 & -(\sqrt{5}+1)/4 & 1/2 \\ 1/2 & -(\sqrt{5}-1)/4 & -(\sqrt{5}+1)/4 \end{bmatrix}, \quad \boldsymbol{p}_{26} = d_3 \begin{bmatrix} (\sqrt{5}-1)/4 \\ 1/2 \\ -(\sqrt{5}+1)/4 \end{bmatrix},$$

$$\mathbf{R}_{27} = \begin{bmatrix} 1/2 & -(\sqrt{5}-1)/4 & (\sqrt{5}+1)/4 \\ -(\sqrt{5}+1)/4 & -1/2 & (\sqrt{5}-1)/4 \\ (\sqrt{5}-1)/4 & -(\sqrt{5}+1)/4 & -1/2 \end{bmatrix}, \quad \boldsymbol{p}_{27} = d_3 \begin{bmatrix} (\sqrt{5}+1)/4 \\ (\sqrt{5}-1)/4 \\ -1/2 \end{bmatrix},$$

$$\mathbf{R}_{28} = \begin{bmatrix} -1/2 & -(\sqrt{5}-1)/4 & (\sqrt{5}+1)/4 \\ -(\sqrt{5}+1)/4 & 1/2 & -(\sqrt{5}-1)/4 \\ -(\sqrt{5}-1)/4 & -(\sqrt{5}+1)/4 & -1/2 \end{bmatrix}, \quad \boldsymbol{p}_{28} = d_3 \begin{bmatrix} (\sqrt{5}+1)/4 \\ -(\sqrt{5}-1)/4 \\ -1/2 \end{bmatrix},$$

$$\mathbf{R}_{29} = \begin{bmatrix} -(\sqrt{5}+1)/4 & 1/2 & (\sqrt{5}-1)/4 \\ (\sqrt{5}-1)/4 & (\sqrt{5}+1)/4 & -1/2 \\ -1/2 & -(\sqrt{5}-1)/4 & -(\sqrt{5}+1)/4 \end{bmatrix}, \quad \mathbf{p}_{29} = d_3 \begin{bmatrix} (\sqrt{5}-1)/4 \\ -1/2 \\ -(\sqrt{5}+1)/4 \end{bmatrix},$$

$$\mathbf{R}_{30} = \begin{bmatrix} 0 & 1 & 0 \\ 1 & 0 & 0 \\ 0 & 0 & -1 \end{bmatrix}, \quad \mathbf{p}_{30} = d_3 \begin{bmatrix} 0 \\ 0 \\ -1 \end{bmatrix}. \tag{D1}$$

Subsequently, referring to the constraint graph in Fig. D1, the $294 \times 180$ original constraint matrix $\mathbf{M}_3$ can be derived as

$$\mathbf{M}_3 = \begin{bmatrix} \mathbf{M}_{11} & \mathbf{0}_{6\times 36} & \mathbf{0}_{6\times 36} & \mathbf{0}_{6\times 36} & \mathbf{0}_{6\times 36} \\ \mathbf{0}_{6\times 36} & \mathbf{M}_{22} & \mathbf{0}_{6\times 36} & \mathbf{0}_{6\times 36} & \mathbf{0}_{6\times 36} \\ \mathbf{0}_{6\times 36} & \mathbf{0}_{6\times 36} & \mathbf{M}_{33} & \mathbf{0}_{6\times 36} & \mathbf{0}_{6\times 36} \\ \mathbf{0}_{6\times 36} & \mathbf{0}_{6\times 36} & \mathbf{0}_{6\times 36} & \mathbf{M}_{44} & \mathbf{0}_{6\times 36} \\ \mathbf{0}_{6\times 36} & \mathbf{0}_{6\times 36} & \mathbf{0}_{6\times 36} & \mathbf{0}_{6\times 36} & \mathbf{M}_{55} \\ \mathbf{0}_{6\times 36} & \mathbf{M}_{62} & \mathbf{0}_{6\times 36} & \mathbf{M}_{64} & \mathbf{M}_{65} \\ \mathbf{0}_{6\times 36} & \mathbf{M}_{72} & \mathbf{M}_{73} & \mathbf{M}_{74} & \mathbf{0}_{6\times 36} \\ \mathbf{M}_{81} & \mathbf{M}_{82} & \mathbf{M}_{83} & \mathbf{0}_{7\times 36} & \mathbf{0}_{7\times 36} \end{bmatrix}, \tag{D2}$$

and the corresponding submatrices are

$$\mathbf{M}_{11} = \begin{bmatrix} \mathbf{S}_1 & \mathbf{0}_6 & \mathbf{0}_6 & \mathbf{0}_6 & \mathbf{0}_6 & \mathbf{0}_6 \\ \mathbf{0}_6 & \mathbf{S}_2 & \mathbf{0}_6 & \mathbf{0}_6 & \mathbf{0}_6 & \mathbf{0}_6 \\ \mathbf{0}_6 & \mathbf{0}_6 & \mathbf{S}_3 & \mathbf{0}_6 & \mathbf{0}_6 & \mathbf{0}_6 \\ \mathbf{0}_6 & \mathbf{0}_6 & \mathbf{0}_6 & \mathbf{S}_4 & \mathbf{0}_6 & \mathbf{0}_6 \\ \mathbf{0}_6 & \mathbf{0}_6 & \mathbf{0}_6 & \mathbf{0}_6 & \mathbf{S}_5 & \mathbf{0}_6 \\ \mathbf{0}_6 & \mathbf{0}_6 & \mathbf{0}_6 & \mathbf{0}_6 & \mathbf{0}_6 & \mathbf{S}_6 \end{bmatrix}, \quad \mathbf{M}_{22} = \begin{bmatrix} \mathbf{S}_7 & \mathbf{0}_6 & \mathbf{0}_6 & \mathbf{0}_6 & \mathbf{0}_6 & \mathbf{0}_6 \\ \mathbf{0}_6 & \mathbf{S}_8 & \mathbf{0}_6 & \mathbf{0}_6 & \mathbf{0}_6 & \mathbf{0}_6 \\ \mathbf{0}_6 & \mathbf{0}_6 & \mathbf{S}_9 & \mathbf{0}_6 & \mathbf{0}_6 & \mathbf{0}_6 \\ \mathbf{0}_6 & \mathbf{0}_6 & \mathbf{0}_6 & \mathbf{S}_{10} & \mathbf{0}_6 & \mathbf{0}_6 \\ \mathbf{0}_6 & \mathbf{0}_6 & \mathbf{0}_6 & \mathbf{0}_6 & \mathbf{S}_{11} & \mathbf{0}_6 \\ \mathbf{0}_6 & \mathbf{0}_6 & \mathbf{0}_6 & \mathbf{0}_6 & \mathbf{0}_6 & \mathbf{S}_{12} \end{bmatrix},$$

$$\mathbf{M}_{33} = \begin{bmatrix} \mathbf{S}_{13} & \mathbf{0}_6 & \mathbf{0}_6 & \mathbf{0}_6 & \mathbf{0}_6 & \mathbf{0}_6 \\ \mathbf{0}_6 & \mathbf{S}_{14} & \mathbf{0}_6 & \mathbf{0}_6 & \mathbf{0}_6 & \mathbf{0}_6 \\ \mathbf{0}_6 & \mathbf{0}_6 & \mathbf{S}_{15} & \mathbf{0}_6 & \mathbf{0}_6 & \mathbf{0}_6 \\ \mathbf{0}_6 & \mathbf{0}_6 & \mathbf{0}_6 & \mathbf{S}_{16} & \mathbf{0}_6 & \mathbf{0}_6 \\ \mathbf{0}_6 & \mathbf{0}_6 & \mathbf{0}_6 & \mathbf{0}_6 & \mathbf{S}_{17} & \mathbf{0}_6 \\ \mathbf{0}_6 & \mathbf{0}_6 & \mathbf{0}_6 & \mathbf{0}_6 & \mathbf{0}_6 & \mathbf{S}_{18} \end{bmatrix}, \quad \mathbf{M}_{44} = \begin{bmatrix} \mathbf{S}_{19} & \mathbf{0}_6 & \mathbf{0}_6 & \mathbf{0}_6 & \mathbf{0}_6 & \mathbf{0}_6 \\ \mathbf{0}_6 & \mathbf{S}_{20} & \mathbf{0}_6 & \mathbf{0}_6 & \mathbf{0}_6 & \mathbf{0}_6 \\ \mathbf{0}_6 & \mathbf{0}_6 & \mathbf{S}_{21} & \mathbf{0}_6 & \mathbf{0}_6 & \mathbf{0}_6 \\ \mathbf{0}_6 & \mathbf{0}_6 & \mathbf{0}_6 & \mathbf{S}_{22} & \mathbf{0}_6 & \mathbf{0}_6 \\ \mathbf{0}_6 & \mathbf{0}_6 & \mathbf{0}_6 & \mathbf{0}_6 & \mathbf{S}_{23} & \mathbf{0}_6 \\ \mathbf{0}_6 & \mathbf{0}_6 & \mathbf{0}_6 & \mathbf{0}_6 & \mathbf{0}_6 & \mathbf{S}_{24} \end{bmatrix},$$

$$\mathbf{M}_{44} = \begin{bmatrix} \mathbf{S}_{25} & \mathbf{0}_6 & \mathbf{0}_6 & \mathbf{0}_6 & \mathbf{0}_6 & \mathbf{0}_6 \\ \mathbf{0}_6 & \mathbf{S}_{26} & \mathbf{0}_6 & \mathbf{0}_6 & \mathbf{0}_6 & \mathbf{0}_6 \\ \mathbf{0}_6 & \mathbf{0}_6 & \mathbf{S}_{27} & \mathbf{0}_6 & \mathbf{0}_6 & \mathbf{0}_6 \\ \mathbf{0}_6 & \mathbf{0}_6 & \mathbf{0}_6 & \mathbf{S}_{28} & \mathbf{0}_6 & \mathbf{0}_6 \\ \mathbf{0}_6 & \mathbf{0}_6 & \mathbf{0}_6 & \mathbf{0}_6 & \mathbf{S}_{29} & \mathbf{0}_6 \\ \mathbf{0}_6 & \mathbf{0}_6 & \mathbf{0}_6 & \mathbf{0}_6 & \mathbf{0}_6 & \mathbf{S}_{30} \end{bmatrix}, \mathbf{M}_{62} = \begin{bmatrix} \mathbf{0}_6 & \mathbf{0}_6 & \mathbf{0}_6 & \mathbf{0}_6 & \mathbf{0}_6 & \mathbf{0}_6 \\ \mathbf{0}_6 & \mathbf{0}_6 & \mathbf{0}_6 & \mathbf{0}_6 & -\mathbf{S}''_{11} & \mathbf{0}_6 \\ \mathbf{0}_6 & \mathbf{0}_6 & \mathbf{0}_6 & \mathbf{0}_6 & \mathbf{0}_6 & \mathbf{0}_6 \\ \mathbf{0}_6 & \mathbf{0}_6 & \mathbf{0}_6 & \mathbf{0}_6 & \mathbf{0}_6 & \mathbf{0}_6 \\ \mathbf{0}_6 & \mathbf{0}_6 & \mathbf{0}_6 & \mathbf{0}_6 & \mathbf{0}_6 & \mathbf{0}_6 \\ \mathbf{0}_6 & \mathbf{0}_6 & \mathbf{0}_6 & \mathbf{0}_6 & \mathbf{0}_6 & \mathbf{0}_6 \end{bmatrix},$$

$$\mathbf{M}_{64} = \begin{bmatrix} \mathbf{0}_6 & \mathbf{0}_6 & \mathbf{0}_6 & \mathbf{0}_6 & \mathbf{0}_6 & -\mathbf{S}''_{24} \\ \mathbf{0}_6 & -\mathbf{S}''_{20} & \mathbf{0}_6 & \mathbf{0}_6 & \mathbf{0}_6 & \mathbf{0}_6 \\ \mathbf{0}_6 & \mathbf{0}_6 & \mathbf{0}_6 & \mathbf{0}_6 & \mathbf{0}_6 & \mathbf{0}_6 \\ \mathbf{0}_6 & \mathbf{0}_6 & \mathbf{0}_6 & \mathbf{0}_6 & -\mathbf{S}''_{23} & \mathbf{0}_6 \\ \mathbf{0}_6 & \mathbf{0}_6 & -\mathbf{S}''_{21} & \mathbf{0}_6 & \mathbf{0}_6 & \mathbf{0}_6 \\ \mathbf{0}_6 & \mathbf{0}_6 & \mathbf{0}_6 & -\mathbf{S}''_{22} & \mathbf{0}_6 & \mathbf{0}_6 \end{bmatrix}, \mathbf{M}_{65} = \begin{bmatrix} \mathbf{0}_6 & \mathbf{0}_6 & \mathbf{0}_6 & \mathbf{0}_6 & -\mathbf{S}''_{29} & -\mathbf{S}'_{30} \\ -\mathbf{S}'_{25} & \mathbf{0}_6 & \mathbf{0}_6 & \mathbf{0}_6 & \mathbf{0}_6 & \mathbf{0}_6 \\ -\mathbf{S}''_{25} & -\mathbf{S}''_{26} & \mathbf{0}_6 & \mathbf{0}_6 & \mathbf{0}_6 & -\mathbf{S}''_{30} \\ \mathbf{0}_6 & \mathbf{0}_6 & \mathbf{0}_6 & -\mathbf{S}'_{28} & -\mathbf{S}'_{29} & \mathbf{0}_6 \\ \mathbf{0}_6 & -\mathbf{S}'_{26} & -\mathbf{S}''_{27} & \mathbf{0}_6 & \mathbf{0}_6 & \mathbf{0}_6 \\ \mathbf{0}_6 & \mathbf{0}_6 & -\mathbf{S}'_{27} & -\mathbf{S}''_{28} & \mathbf{0}_6 & \mathbf{0}_6 \end{bmatrix},$$

$$\mathbf{M}_{72} = \begin{bmatrix} \mathbf{0}_6 & \mathbf{0}_6 & \mathbf{0}_6 & -\mathbf{S}''_{10} & \mathbf{0}_6 & \mathbf{0}_6 \\ \mathbf{0}_6 & \mathbf{0}_6 & \mathbf{0}_6 & \mathbf{0}_6 & \mathbf{0}_6 & \mathbf{0}_6 \\ \mathbf{0}_6 & \mathbf{0}_6 & \mathbf{0}_6 & \mathbf{0}_6 & \mathbf{0}_6 & \mathbf{0}_6 \\ \mathbf{0}_6 & \mathbf{0}_6 & -\mathbf{S}''_{9} & \mathbf{0}_6 & \mathbf{0}_6 & \mathbf{0}_6 \\ \mathbf{0}_6 & \mathbf{0}_6 & \mathbf{0}_6 & \mathbf{0}_6 & \mathbf{0}_6 & \mathbf{0}_6 \\ \mathbf{0}_6 & \mathbf{0}_6 & \mathbf{0}_6 & \mathbf{0}_6 & \mathbf{0}_6 & -\mathbf{S}''_{12} \end{bmatrix}, \mathbf{M}_{73} = \begin{bmatrix} \mathbf{0}_6 & \mathbf{0}_6 & \mathbf{0}_6 & \mathbf{0}_6 & \mathbf{0}_6 & \mathbf{0}_6 \\ \mathbf{0}_6 & \mathbf{0}_6 & \mathbf{0}_6 & \mathbf{0}_6 & \mathbf{0}_6 & -\mathbf{S}''_{18} \\ \mathbf{0}_6 & \mathbf{0}_6 & \mathbf{0}_6 & -\mathbf{S}''_{16} & -\mathbf{S}''_{17} & \mathbf{0}_6 \\ \mathbf{0}_6 & \mathbf{0}_6 & \mathbf{0}_6 & \mathbf{0}_6 & -\mathbf{S}'_{17} & -\mathbf{S}'_{18} \\ \mathbf{0}_6 & -\mathbf{S}''_{14} & -\mathbf{S}''_{15} & \mathbf{0}_6 & \mathbf{0}_6 & \mathbf{0}_6 \\ -\mathbf{S}''_{13} & \mathbf{0}_6 & \mathbf{0}_6 & \mathbf{0}_6 & \mathbf{0}_6 & \mathbf{0}_6 \end{bmatrix},$$

$$\mathbf{M}_{74} = \begin{bmatrix} -\mathbf{S}'_{19} & -\mathbf{S}'_{20} & \mathbf{0}_6 & \mathbf{0}_6 & \mathbf{0}_6 & \mathbf{0}_6 \\ -\mathbf{S}''_{19} & \mathbf{0}_6 & \mathbf{0}_6 & \mathbf{0}_6 & \mathbf{0}_6 & -\mathbf{S}'_{24} \\ \mathbf{0}_6 & \mathbf{0}_6 & \mathbf{0}_6 & \mathbf{0}_6 & -\mathbf{S}'_{23} & \mathbf{0}_6 \\ \mathbf{0}_6 & \mathbf{0}_6 & \mathbf{0}_6 & \mathbf{0}_6 & \mathbf{0}_6 & \mathbf{0}_6 \\ \mathbf{0}_6 & \mathbf{0}_6 & \mathbf{0}_6 & -\mathbf{S}'_{22} & \mathbf{0}_6 & \mathbf{0}_6 \\ \mathbf{0}_6 & \mathbf{0}_6 & -\mathbf{S}'_{21} & \mathbf{0}_6 & \mathbf{0}_6 & \mathbf{0}_6 \end{bmatrix}, \mathbf{M}_{81} = \begin{bmatrix} \mathbf{0}_6 & \mathbf{0}_6 & -\mathbf{S}''_{3} & -\mathbf{S}'_{4} & \mathbf{0}_6 & \mathbf{0}_6 \\ \mathbf{0}_6 & \mathbf{0}_6 & \mathbf{0}_6 & -\mathbf{S}''_{4} & -\mathbf{S}'_{5} & \mathbf{0}_6 \\ \mathbf{0}_6 & \mathbf{0}_6 & \mathbf{0}_6 & \mathbf{0}_6 & \mathbf{0}_6 & -\mathbf{S}''_{6} \\ \mathbf{0}_6 & \mathbf{0}_6 & \mathbf{0}_6 & \mathbf{0}_6 & \mathbf{0}_6 & \mathbf{0}_6 \\ \mathbf{0}_6 & \mathbf{0}_6 & \mathbf{0}_6 & \mathbf{0}_6 & \mathbf{0}_6 & \mathbf{0}_6 \\ \mathbf{0}_6 & -\mathbf{S}''_{2} & -\mathbf{S}'_{3} & \mathbf{0}_6 & \mathbf{0}_6 & \mathbf{0}_6 \\ -\mathbf{S}'_{1} & \mathbf{0}_6 & \mathbf{0}_6 & \mathbf{0}_6 & -\mathbf{S}''_{5} & -\mathbf{S}'_{6} \end{bmatrix},$$

$$\mathbf{M}_{82} = \begin{bmatrix} \mathbf{0}_6 & \mathbf{0}_6 & -\mathbf{S}'_{9} & \mathbf{0}_6 & \mathbf{0}_6 & \mathbf{0}_6 \\ \mathbf{0}_6 & \mathbf{0}_6 & \mathbf{0}_6 & -\mathbf{S}'_{10} & \mathbf{0}_6 & \mathbf{0}_6 \\ \mathbf{0}_6 & \mathbf{0}_6 & \mathbf{0}_6 & \mathbf{0}_6 & -\mathbf{S}'_{11} & -\mathbf{S}'_{12} \\ \mathbf{0}_6 & -\mathbf{S}''_{8} & \mathbf{0}_6 & \mathbf{0}_6 & \mathbf{0}_6 & \mathbf{0}_6 \\ -\mathbf{S}''_{7} & \mathbf{0}_6 & \mathbf{0}_6 & \mathbf{0}_6 & \mathbf{0}_6 & \mathbf{0}_6 \\ \mathbf{0}_6 & -\mathbf{S}'_{8} & \mathbf{0}_6 & \mathbf{0}_6 & \mathbf{0}_6 & \mathbf{0}_6 \\ \mathbf{0}_6 & \mathbf{0}_6 & \mathbf{0}_6 & \mathbf{0}_6 & \mathbf{0}_6 & \mathbf{0}_6 \end{bmatrix}, \mathbf{M}_{83} = \begin{bmatrix} \mathbf{0}_6 & \mathbf{0}_6 & \mathbf{0}_6 & \mathbf{0}_6 & \mathbf{0}_6 & \mathbf{0}_6 \\ \mathbf{0}_6 & \mathbf{0}_6 & \mathbf{0}_6 & \mathbf{0}_6 & \mathbf{0}_6 & \mathbf{0}_6 \\ \mathbf{0}_6 & \mathbf{0}_6 & \mathbf{0}_6 & \mathbf{0}_6 & \mathbf{0}_6 & \mathbf{0}_6 \\ \mathbf{0}_6 & \mathbf{0}_6 & -\mathbf{S}'_{15} & -\mathbf{S}'_{16} & \mathbf{0}_6 & \mathbf{0}_6 \\ -\mathbf{S}'_{13} & -\mathbf{S}'_{14} & \mathbf{0}_6 & \mathbf{0}_6 & \mathbf{0}_6 & \mathbf{0}_6 \\ \mathbf{0}_6 & \mathbf{0}_6 & \mathbf{0}_6 & \mathbf{0}_6 & \mathbf{0}_6 & \mathbf{0}_6 \\ \mathbf{0}_6 & \mathbf{0}_6 & \mathbf{0}_6 & \mathbf{0}_6 & \mathbf{0}_6 & \mathbf{0}_6 \end{bmatrix}.$$

Finally, the rank of the original constraint matrix $\mathbf{M}_3$ is 179, the conclusion that the deployable dodecahedral mechanism has one mobility can be generated as $m = n - rank(\mathbf{M}_3) = 180 - 179 = 1$.